%% file: main.tex
\documentclass[%
    letterpaper,           
    colorlinks,            
    font = lmodern,         
    palette = gt,           
    sectionstyle = runin,   
    headingcolor = palette, 
    fontsize = 11pt,        
]{tkneurips}

\usepackage[english]{babel}
\usepackage{graphicx}
\usepackage{mathtools}
\usepackage{amsopn}
\usepackage[acronym]{glossaries}
\setkeys{glslink}{hyper=false}
\usepackage{bm}
\usepackage{booktabs}
\usepackage{multirow}
\usepackage{enumitem}
\setlist[itemize]{leftmargin=*}
\usepackage{subcaption}
\usepackage{custom}  
\usepackage{placeins}

\theoremstyle{definition}

\theoremstyle{remark}

\begin{document}

\title{\Large Greedy Orthogonal Polynomial Manifolds with Lipschitz Control for Dimensionality Reduction and Reduced Modeling\thanks{\textbf{Funding}: The
        authors TK and EQ were supported in part by the Department of Energy Office of
        Science Advanced Scientific Computing Research, DOE Award
        DE-SC0024721.}}

\author{%
    Tomoki Koike  \\
    School of Aerospace Engineering, 
    Georgia Institute of Technology, Atlanta, GA, USA \\
    \email{tkoike@gatech.edu}
    \And
    Benjamin Peherstorfer \\
    Institute of Mathematics, EPFL, 
    Lausanne, Switzerland \\
    \email{benjamin.peherstorfer@epfl.ch}
    \And
    Elizabeth Qian \\
    School of Aerospace Engineering and School of Computational Science and Engineering \\
    Georgia Institute of Technology, Atlanta, GA, USA \\
    \email{eqian@gatech.edu}
}

\shorttitle{Greedy Orthogonal Polynomial Manifold}
\shortauthor{T. Koike et al.}
\shortdate{}              
\shortinstitute{}        


\abstract{\small%
    Polynomial manifolds provide nonlinear low-dimensional representations through polynomial decoders. We propose to represent the polynomial terms of the decoder using features that are orthonormal on the training data. This empirical orthogonality enables a greedy algorithm, Greedy-OPM, which jointly selects encoder modes and polynomial features while minimizing the reconstruction error under a prescribed bound on the decoder's Lipschitz constant. Numerical experiments show that this Lipschitz constraint improves generalization away from the training data while retaining high reconstruction accuracy.
}

\maketitle

\input{sections/intro}

\input{sections/background}
\input{sections/method}

\input{sections/numerics}

\input{sections/conclusion}

\appendix
\input{sections/extra_p5_plots}

\bibliographystyle{plainurl}
\bibliography{refs}

\end{document}

%% file: sections/intro.tex
\section{Introduction}\label{sec:intro}
Polynomial manifolds provide nonlinear low-dimensional approximations in which low-dimensional vectors are mapped to the high-dimensional ambient space by a polynomial decoder and high-dimensional vectors are encoded with linear encoders. It is well recognized that optimizing an encoder-decoder pair for reconstruction accuracy alone need not yield a stable nonlinear approximation  and that additional control of its Lipschitz constant is required~\cite{cohen2022Optimal,bensalah2026Nonlinear}. A related empirical observation has been made for polynomial manifolds in model reduction, where minimizing the reconstruction error alone can produce nonsmooth reduced coordinates~\cite{schwerdtner2026Operator}.

We propose the \gls{greedyopm} approach, a greedy construction of encoder-decoder pairs of polynomial
manifolds that accounts explicitly for both reconstruction accuracy and sensitivity to perturbations. The key novelty is to represent the polynomial feature map in a basis that is orthonormal with respect to the empirical inner product on the reduced coordinates.
Orthonormality improves the conditioning of the decoder fit, but more importantly it changes the structure of the problem of selecting encoder modes and polynomial features.
Namely, the polynomial features decouple on the training data, so that their contributions to the reconstruction error and to a bound on the Lipschitz constant of the decoder can be assessed individually. This decoupling makes it possible to impose a constraint on the Lipschitz constant while minimizing the reconstruction error.
Moreover, the orthogonal representation enables an efficient greedy construction because the polynomial fit and the effect on the Lipschitz constant bound of selecting or suppressing individual polynomial features can be evaluated without solving coupled regression problems.

Quadratic and more general polynomial manifolds have been applied to nonlinear structural dynamics~\cite{jain2017Quadratic,rutzmoser2017Generalization}, projection-based reduction of transport-dominated systems~\cite{barnett2022Quadratic,cohen2023Nonlinear}, nonintrusive reduced modeling~\cite{geelen2023Operator,benner2023QuadraticDecoder}, data-driven polynomial representation learning~\cite{geelen2023Learning,geelen2024Learning}, structure-preserving reduction of Hamiltonian systems~\cite{sharma2023Symplectic}, reconstruction from sparse measurements~\cite{schwerdtner2025SparseRegression}, online learning from streaming data~\cite{schwerdtner2025Online}, Neural Galerkin methods~\cite{weder2025NeuralGalerkin}, and neural network manifolds~\cite{koike2026Sparse,barnett2023Neural,aresdeparga2026Nonlinear}. Approximation bound for polynomially mapped manifolds has been established in~\cite{buchfink2024Approximation}.

Different methods have been proposed for constructing polynomial manifolds. The works~\cite{jain2017Quadratic,rutzmoser2017Generalization} construct quadratic manifolds for nonlinear structural dynamics from vibration modes and modal derivatives so that the curvature of the manifold is informed directly by the governing mechanical system. In contrast, the works~\cite{barnett2022Quadratic,geelen2023Operator} use \gls{pod} to define a linear encoder and learn the quadratic polynomial decoder via regression from data. The approach~\cite{geelen2023Operator} additionally uses group-sparse regularization to select quadratic features. The works~\cite{geelen2023Learning,geelen2024Learning} formulate a joint optimization of the encoder and polynomial decoder and solve the resulting problem with alternating optimization methods. The approach~\cite{paxton2026Fast} instead formulates quadratic-manifold learning as a continuous optimization problem over orthonormal encoder bases on the Stiefel manifold. A system-theoretic construction is proposed in~\cite{padhi2026beyond}, where the quadratic projection matrices are determined directly through nonlinear moment matching rather than through data-fitting optimization.

A building block for our work is the greedy algorithm introduced in~\cite{schwerdtner2026Greedy}. Its key observation is that the leading \gls{pod} directions, while optimal for linear reconstruction, need not provide an encoder that is most effective for a polynomial feature map. The method therefore selects encoder modes from a larger pool of left singular vectors according to the reconstruction error attainable by the resulting polynomial encoder-decoder pair.
The work~\cite{bensalah2026Nonlinear} addresses the  construction of accurate polynomial manifolds with a focus on controlled Lipschitz stability; however, the authors of~\cite{bensalah2026Nonlinear} use compositional networks with local nodes represented by polynomials that do not lead to a global low-degree polynomial structure. The decoder is built as a tree-structured composition of polynomial maps, while the encoder coordinates and the dependencies of the component maps are selected adaptively based on approximation and stability requirements.

Our approach follows a different construction. For each candidate encoder mode, we determine the polynomial decoder that minimizes the reconstruction error subject to the prescribed Lipschitz constraint.
Because of the empirical orthogonality, this constrained problem reduces to a projection problem that has a closed-form group soft-thresholding solution. Encoder modes and nonlinear features are therefore selected together, with accuracy and Lipschitz control handled within a single greedy criterion rather than by a general constrained solve for every candidate encoder direction.
We construct the empirical orthogonal polynomial basis directly on the reduced training coordinates using multivariate Vandermonde-Arnoldi (V+A) orthogonalization~\cite{brubeck2021Vandermonde,austin2021Multivariate,hokanson2020Multivariate,zhu2025Convergence}.

On transport-dominated parametric problems, we demonstrate that our \gls{greedyopm} with the Lipschitz constraint improves stability and extrapolation  without sacrificing the reconstruction accuracy gained from nonlinear manifold approximation.

The remainder of this paper is organized as follows. In \Cref{sec:bg}, we review polynomial manifolds and their greedy construction from~\cite{schwerdtner2026Greedy}. In \Cref{sec:opm}, we introduce the orthogonal polynomial manifold representation, and, in \Cref{sec:lipschitz}, we present \gls{greedyopm} with its Lipschitz control strategy. In \Cref{sec:numerics}, we demonstrate \gls{greedyopm} on numerical experiments, and we conclude in \Cref{sec:conclusion}.

%% file: sections/background.tex
\section{Preliminaries}\label{sec:bg}
In this section, we provide preliminaries on greedily constructing encoder-decoder pairs of polynomial manifolds. 

\subsection{Polynomial encoder-decoder models}\label{sec:bg:manifold}
Let $\sbold^{(1)},\ldots,\sbold^{(M)}\in\R^{N}$ denote $M$ data points of dimension $N$. Given a reference state $\sref\in\R^N$, for example the data mean, we define the (shifted) data matrix as
    \begin{equation}\label{eq:shifted_data}
       \Sbold
    =
    [\sbold^{(1)}-\sref,\ldots,\sbold^{(M)}-\sref]
    \in\R^{N\times M}.
\end{equation}
We approximate the data through an encoder-decoder pair. The encoder
is defined by an orthonormal basis matrix 
$\Vbold=[\vbold_1,\ldots,\vbold_n]\in\R^{N\times n}$ with
$\Vbold^\top\Vbold=\Id_n$ and is given by
\begin{equation}\label{eq:encoder}
    \Phi:\R^N\rightarrow\R^n,
    \qquad
    \Phi(\sbold)
    =
    \Vbold^\top(\sbold-\sref).
\end{equation}
We denote the corresponding affine reconstruction map by
\begin{equation}\label{eq:encoder_dagger}
    \Phi^\dagger:\R^n\rightarrow\R^N,
    \qquad
    \Phi^\dagger(\shat)
    =
    \sref+\Vbold\shat.
\end{equation}
Since $\Vbold^\top\Vbold=\Id_n$, we have
$\Phi(\Phi^\dagger(\shat))=\shat$ for every $\shat\in\R^n$.

The decoder is a polynomial function,
\begin{equation}\label{eq:poly_manifold}
    \Gamma(\shat)
    =
    \Phi^\dagger(\shat)
    +
    \Vboldbar\,\pbold(\shat),
\end{equation}
with a given polynomial feature map of degree between $2$ and $p$, for a fixed integer $p \geq 2$,
\begin{equation}\label{eq:Prelim:PolyFeatureMap}
    \pbold:\R^n\rightarrow\R^{m_n}\,,
\end{equation}
and coefficient matrix $\Vboldbar\in\R^{N\times m_n}$. The number of polynomial features is  $m_n=\binom{n+p}{p}-(n+1)$.

Evaluating the polynomial feature map at the reduced data vectors, 
\[
\shat^{(i)} = \Phi(\sbold^{(i)})\,,\qquad i = 1, \dots, M\,,
\]
gives
\[
    \Pbold
    =
    [\pbold(\shat^{(1)}),\ldots,\pbold(\shat^{(M)})]
    \in\R^{m_n\times M}.
\]

\subsection{Greedy selection of the reduced modes}\label{sec:bg:greedy}
To train a linear encoder and polynomial decoder pair, the reduced basis $\Vbold$ and polynomial coefficients $\Vboldbar$ need to be chosen.
Early works~\cite{geelen2023Operator,barnett2022Quadratic} defined the reduced basis $\Vbold$ using the leading principal components of $\Sbold$. However, while this choice is optimal for the linear reconstruction map \eqref{eq:encoder_dagger}, it can be a poor choice for polynomial reconstruction~\cite{cohen2023Nonlinear,schwerdtner2026Greedy}.

The work \cite{schwerdtner2026Greedy} proposes to greedily select the basis vectors forming the matrix $\Vbold$ according to the reconstruction error of the linear encoder and polynomial decoder pair from the first $s > n$ left-singular vectors of $\Sbold$.
Let $\Ubold\Sigmasmat\Wbold^\top$ be a rank-$s$ truncated \gls{svd} of $\Sbold$ with $s>n$, where $\Ubold=[\ubold_1,\ldots,\ubold_s]\in\R^{N\times s}$ contains the candidate basis vectors, $\Sigmasmat=\diag(\sigma_1,\ldots,\sigma_s)$, and $\Wbold\in\R^{M\times s}$. The $n$ columns of $\Vbold$ are selected from the $s$ columns $\ubold_1,\dots,\ubold_s$ of $\Ubold$.
Suppose that after $i-1$ greedy steps the current basis is
$\Vbold_{i-1}=[\vbold_1,\ldots,\vbold_{i-1}]$ and a new basis vector
$\vbold_i$ is to be selected from the remaining candidates. For a
previously unselected candidate vector $\ubold_j$, define the candidate
basis
\begin{equation}\label{eq:candidate_basis}
    \Vbold_{i-1}^{+j}
    =
    [\Vbold_{i-1},\ubold_j]
\end{equation}
and the corresponding candidate encoder
\begin{equation}\label{eq:candidate_encoder}
    \Phi^{+j}(\sbold)
    =
    (\Vbold_{i-1}^{+j})^\top(\sbold-\sref).
\end{equation}
Its corresponding reconstruction map is
\[
    (\Phi^{+j})^\dagger(\shat)
    =
    \sref+\Vbold_{i-1}^{+j}\shat.
\]
The reduced coordinates and projection error are
\begin{equation}\label{eq:candidate_reduced_coordinates}
    \Shat^{+j}
    =
    [\Phi^{+j}(\sbold^{(1)}),\ldots,\Phi^{+j}(\sbold^{(M)})]
    =
    (\Vbold_{i-1}^{+j})^\top\Sbold,
    \qquad
    \Ebold^{+j}
    =
    \Sbold-\Vbold_{i-1}^{+j}\Shat^{+j}.
\end{equation}
Furthermore, let $\Pbold^{+j}$ denote the polynomial feature matrix obtained by evaluating $\pbold$ at the columns of $\Shat^{+j}$.

The quality of candidate vector $\ubold_j$ is measured by the smallest reconstruction error attainable with the corresponding candidate basis, which leads to the objective
\begin{align}
    J_i(j)
    &=
    \min_{\Vboldbar\in\mathbb{R}^{N\times m_n}}
    \frac12
    \big\|
        \Ebold^{+j}
        -
        \Vboldbar\Pbold^{+j}
    \big\|_\frob^2.
    \label{eq:greedy_objective}
\end{align}
The greedy approach selects, among all previously unselected candidate vectors $\ubold_j$, one that minimizes~\eqref{eq:greedy_objective}. The selected vector is denoted by $\vbold_i$, and the basis is updated as $\Vbold_i=[\Vbold_{i-1},\vbold_i]$.

This procedure is repeated until $n$ basis vectors have been selected, yielding the basis matrix $\Vbold=[\vbold_1,\ldots,\vbold_n]$.
The coefficient matrix $\Vboldbar$ is then determined by
\begin{equation}\label{eq:poly_manifold_ls}
    \Vboldbar
    =
    \argmin_{\Vboldbar\in\mathbb{R}^{N\times m_n}}
    \frac12
    \left\|
        \Ebold-\Vboldbar\Pbold
    \right\|_\frob^2\,,
\end{equation}
with $\Ebold = \Sbold - \Vbold\Vbold^\top\Sbold$. 

The least-squares problem~\eqref{eq:poly_manifold_ls} is typically
ill-conditioned, since the monomial features in \(\Pbold\) are nearly linearly
dependent on the data and the condition number of \(\Pbold\Pbold^\top\) grows
rapidly with the degree \(p\) and the reduced dimension \(n\). Moreover, the
number of features \(m_n\) grows combinatorially in \(p\) and \(n\), while
typically only a small subset contributes to the reconstruction, so that the
fitted decoder tends to overfit the training data and generalizes poorly.

%% file: sections/method.tex
\section{Orthogonal polynomial manifolds}\label{sec:opm}
We propose to use an orthonormal representation of the polynomial features. 

\subsection{Degree-based decomposition of polynomial spaces}\label{sec:DegreeBasedDecomp}
Let \(\mathcal{P}_p(\R^n)\) denote the space of scalar-valued polynomials defined over \(\R^n\) of total degree at most \(p\). The space \(\mathcal{P}_p(\R^n)\) can be decomposed according to the polynomial degree into the subspace \(\mathcal{P}_1(\R^n)\) spanned by the constant and degree-one polynomials and the rest \(\mathcal{R}_p(\R^n)\), which contains polynomials of degree two and higher,
    \begin{equation}\label{eq:poly_space_split}
        \mathcal{P}_p(\R^n)
        =
        \mathcal{P}_1(\R^n)
        \oplus
        \mathcal{R}_p(\R^n)\,.
    \end{equation}
The subspace \(\mathcal{P}_1(\R^n)\) has dimension \(n+1\) and \(\mathcal{R}_p(\R^n)\) has dimension \(m_n\).
For example, for \(n=3\) and \(p=2\), the two subspaces can be represented using the monomial basis as
\[
    \mathcal{P}_2(\R^3)
    =
    \underbrace{
        \operatorname{span}\{1,x_1,x_2,x_3\}
    }_{\mathcal{P}_1(\R^3)}
    \; \oplus \;
    \underbrace{
        \operatorname{span}\{
            x_1^2,
            x_1x_2,
            x_1x_3,
            x_2^2,
            x_2x_3,
            x_3^2
        \}
    }_{\mathcal{R}_2(\R^3)}.
\]
The decomposition~\eqref{eq:poly_space_split} is reflected directly in the polynomial decoder~\eqref{eq:poly_manifold}. The affine term
\(\Phi^\dagger(\shat)\) belongs to \(\mathcal{P}_1(\R^n)\), while the nonlinear correction
\(\Vboldbar\pbold(\shat)\) is represented in \(\mathcal{R}_p(\R^n)\) so that every component \(\Gamma_1,\dots,\Gamma_N:\R^n\to\R\) of
\(\Gamma(\shat)=[\Gamma_1(\shat),\dots,\Gamma_N(\shat)]\tran \)
can be decomposed into the sum of a function in
\(\mathcal{P}_1(\R^n)\) and \(\mathcal{R}_p(\R^n)\),
\[
    \Gamma_i(\shat)
    =
    \underbrace{[\Phi^\dagger(\shat)]_i}_{\mathcal{P}_1(\R^n)}
    \; + \;
    \underbrace{\Gamma_i^{\mathrm{nonlin}}(\shat)}_{\mathcal{R}_p(\R^n)},
    \qquad i=1,\dots,N.
\]
In vector form, we have
\begin{equation}\label{eq:OMP:NonOrthoGammaRep}
    \Gamma(\shat)
    =
    \Phi^\dagger(\shat)
    +
    \Gamma^{\mathrm{nonlin}}(\shat)
    =
    \underbrace{\Phi^\dagger(\shat)}_{\mathcal{P}_1(\R^n)}
    \; + \;
    \underbrace{\Vboldbar\pbold(\shat)}_{\mathcal{R}_p(\R^n)}.
\end{equation}

\subsection{Orthogonal decompositions}\label{sec:opm:orthopoly}
In the following, we will seek a representation of \(\Gamma\) that decomposes into the affine terms and the orthogonal complement with respect to a given inner product $\langle \cdot,\cdot \rangle$. We note that the degree-based decomposition~\eqref{eq:OMP:NonOrthoGammaRep} is orthogonal with respect to the coefficient inner product, but we will be interested in another inner product later.

 Let \(\Pi_1\) denote the orthogonal projection onto
\(\mathcal{P}_1(\R^n)\) with respect to the inner product $\langle \cdot, \cdot \rangle$. We define the projected space
\begin{displaymath}
    \mathcal{R}_p^\perp(\R^n)
    =
    \left\{
        q-\Pi_1q
        \;:\;
        q\in\mathcal{R}_p(\R^n)
    \right\},
\end{displaymath}
so that every element of \(\mathcal{R}_p^\perp(\R^n)\) is 
orthogonal to every element of \(\mathcal{P}_1(\R^n)\). Note that the
space \(\mathcal{R}_p^\perp(\R^n)\) has the same dimension \(m_n\) as
\(\mathcal{R}_p(\R^n)\), but it is in general a different subspace of
\(\mathcal{P}_p(\R^n)\).
Correspondingly, we obtain the orthogonal decomposition
\begin{displaymath}
    \mathcal{P}_p(\R^n)
    =
    \mathcal{P}_1(\R^n)
    \oplus
    \mathcal{R}_p^\perp(\R^n),
\end{displaymath}
into the subspace $\mathcal{P}_1(\R^n)$ corresponding to the affine terms and the orthogonal complement $\mathcal{R}_p^\perp(\R^n)$. 
For the coefficient inner product, the orthogonal complement $\mathcal{R}_p^{\perp}(\mathbb{R}^n)$ coincides with $\mathcal{R}_p(\mathbb{R}^n)$, and we recover the degree-based decomposition of \Cref{sec:DegreeBasedDecomp}. 

Consider now an orthonormal basis
    \(\phi_0,\ldots,\phi_n\) of \(\mathcal{P}_1(\R^n)\) and an 
    orthonormal basis \(q_1,\ldots,q_{m_n}\) of
    \(\mathcal{R}_p^\perp(\R^n)\). Since the two subspaces are orthogonal,
    the union of the sets of basis functions forms an 
    orthonormal basis of \(\mathcal{P}_p(\R^n)\).
For a given linear encoder \(\Phi: \mathbb{R}^N \to \mathbb{R}^n\), we now use the orthogonal decomposition and the corresponding orthogonal bases to represent a decoder as
\begin{equation}\label{eq:GammaBRep}
    \Gamma(\shat)
    =
    \Phi^\dagger(\shat)
    +
    \Bboldbar\qbold(\shat),
\end{equation}
where each component function of the affine decoder \(\Phi^{\dagger}\) is in \(\mathcal{P}_1(\mathbb{R}^n)\) and 
the nonlinear polynomial feature map is represented in the basis
\(q_1,\ldots,q_{m_n}\) of \(\mathcal{R}_p^\perp(\R^n)\) as
\begin{displaymath}
    \qbold(\shat)
    =
    [q_1(\shat),\ldots,q_{m_n}(\shat)]\tran
    \in\R^{m_n}.
\end{displaymath}

\subsection{Greedy mode selection with orthogonal polynomial features}
\label{sec:opm:greedy}
We now restate the greedy method from \Cref{sec:bg:greedy} using the orthogonal decoder representation~\eqref{eq:GammaBRep}. To this end, we choose the inner product that defines the orthogonal polynomial features based on the training data.
   
\subsubsection{Empirical inner product}
We consider orthogonality with respect to the  empirical inner product
\begin{equation}\label{eq:emp_ip}
    \langle f,g\rangle_M
    =
    \frac{1}{M}
    \sum_{\ell=1}^{M}
    f(\shat^{(\ell)})g(\shat^{(\ell)})\,,
\end{equation}
where $f,g\in\mathcal{P}_p(\R^n)$. We assume, for simplicity, that the reduced data points are unisolvent for $\mathcal{P}_p(\R^n)$, which requires $M\geq\binom{n+p}{n}$ and ensures that \eqref{eq:emp_ip} defines an inner product on $\mathcal{P}_p(\R^n)$. This assumption is not essential for the subsequent greedy procedure, however, since the least-squares computations rely only on the polynomial feature evaluations at the training data points.

Evaluating the polynomial feature map at the reduced data points gives
\[
    \Qbold
    =
    [\qbold(\shat^{(1)}),\ldots,\qbold(\shat^{(M)})]
    \in\R^{m_n\times M}.
\]
Empirical orthonormality and the orthogonality between
\(\mathcal{P}_1(\R^n)\) and \(\mathcal{R}_p^\perp(\R^n)\) then imply
\begin{equation}\label{eq:Q_orthogonality}
    \frac{1}{M}\Qbold\Qbold^\top
    =
    \Id_{m_n},
    \qquad
    \Qbold\mathbf{1}_M
    =
    \0,
    \qquad
    \Qbold\Shat^\top
    =
    \0.
\end{equation}
Thus, the nonlinear features are mutually orthonormal on the data and
are also orthogonal to both the constant and linear components.

\subsubsection{Greedy steps}
As in the greedy procedure in \Cref{sec:bg:greedy}, we seek a basis \(\Vbold\) for the 
encoder \(\Phi\) but now use an orthogonal representation of the corresponding decoder \(\Gamma\) during the greedy iteration.
Suppose that after \(i-1\)
greedy steps the current encoder basis is \(\Vbold_{i-1}\). For an
unselected candidate basis vector \(\ubold_j\), we form \( \Vbold_{i-1}^{+j} \) as in~\eqref{eq:candidate_basis}
with corresponding candidate encoder \( \Phi^{+j}(\sbold) \) 
and reduced coordinates \( \Shat^{+j} \) from~\eqref{eq:candidate_encoder} and~\eqref{eq:candidate_reduced_coordinates}, respectively.

For the reduced coordinates \(\Shat^{+j}\), we construct the
corresponding empirical orthonormal nonlinear feature matrix
\(\Qbold^{+j}\). To this end, we order the monomials of total degree at most
\(p\) with the constant and linear monomials first, and apply the multivariate
Vandermonde-Arnoldi procedure
\cite{brubeck2021Vandermonde,austin2021Multivariate,
hokanson2020Multivariate,zhu2025Convergence}
to their evaluations on the columns of \(\Shat^{+j}\). The procedure
orthonormalizes these polynomials with respect to the empirical inner
product~\eqref{eq:emp_ip}; after the constant and linear components, the
remaining \(m_i\) polynomials form an empirical orthonormal basis of
\(\mathcal{R}_p^\perp(\R^i)\). Evaluating these nonlinear basis functions on
the columns of \(\Shat^{+j}\) gives \(\Qbold^{+j}\), which satisfies
\begin{equation}\label{eq:candidate_Q_orthogonality}
    \frac{1}{M}
    \Qbold^{+j}(\Qbold^{+j})^\top
    =
    \Id_{m_i},
    \qquad
    \Qbold^{+j}\mathbf{1}_M=\0,
    \qquad
    \Qbold^{+j}(\Shat^{+j})^\top=\0.
\end{equation}
The corresponding decoder is represented as
    \[
        \Gamma^{+j}(\shat)
        =
        (\Phi^{+j})^\dagger(\shat)
        +
        \Bboldbar^{+j}\qbold^{+j}(\shat),
    \]
    where
    \[
        (\Phi^{+j})^\dagger(\shat)
        =
        \sref+\Vbold_{i-1}^{+j}\shat.
    \]
The projection error of the affine reconstruction is
    \[
        \Ebold^{+j}
        =
        \Sbold-\Vbold_{i-1}^{+j}\hat{\Sbold}^{+j}.
    \]
  
  Analogous to the greedy formulation in \Cref{sec:bg:greedy}, the quality of candidate basis vector \(\ubold_j\) is measured by the
    corresponding orthogonal polynomial reconstruction error
    \begin{equation}\label{eq:greedy_objective_perp}
        J_i^\perp(j)
        =
        \min_{\Bboldbar^{+j}}
        \frac12
        \left\|
            \Ebold^{+j}
            -
            \Bboldbar^{+j}\Qbold^{+j}
        \right\|_\frob^2.
    \end{equation}
The identities~\eqref{eq:candidate_Q_orthogonality} simplify the objective of the 
least-squares problem~\eqref{eq:greedy_objective_perp} as 
\begin{equation}
\begin{aligned}
    \frac12\left\|
        \Ebold^{+j}
        -
        \Bboldbar^{+j}\Qbold^{+j}
    \right\|_\frob^2
    &=\frac12\|\Ebold^{+j}\|_\frob^2
        \; - \; \\
    & \qquad 
        \operatorname{tr}
        \left(
            \Bboldbar^{+j}
            \Qbold^{+j}
            (\Ebold^{+j})^\top
        \right)
        +
        \frac12
        \operatorname{tr}
        \left(
            \Bboldbar^{+j}
            \Qbold^{+j}
            (\Qbold^{+j})^\top
            (\Bboldbar^{+j})^\top
        \right)\,.
\end{aligned}
\end{equation}
Using~\eqref{eq:candidate_Q_orthogonality}, we obtain 
\[
    J_i^\perp(j)
    =
    \min_{\Bboldbar^{+j}}
    \left[
        \frac12\|\Ebold^{+j}\|_\frob^2
        -
        \operatorname{tr}
        \left(
            \Bboldbar^{+j}
            \Qbold^{+j}
            (\Ebold^{+j})^\top
        \right)
        +
        \frac{M}{2}
        \|\Bboldbar^{+j}\|_\frob^2
    \right].
\]
The first-order optimality condition is therefore
\[
    M\Bboldbar^{+j}
    -
    \Ebold^{+j}(\Qbold^{+j})^\top
    =
    \0,
\]
and hence
\begin{equation}\label{eq:candidate_Bbar_solution}
    \Bboldbar^{+j}
    =
    \frac{1}{M}
    \Ebold^{+j}(\Qbold^{+j})^\top.
\end{equation}
Substituting this solution back into the objective yields
\begin{equation}\label{eq:greedy_objective_perp_closed}
    J_i^\perp(j)
    =
    \frac12\|\Ebold^{+j}\|_\frob^2
    -
    \frac{1}{2M}
    \left\|
        \Ebold^{+j}(\Qbold^{+j})^\top
    \right\|_\frob^2.
\end{equation}
Thus, a least-squares solve is no longer necessary, as it is replaced by~\eqref{eq:candidate_Bbar_solution}.

The greedy selection itself is unchanged in form: among all
previously unselected candidate basis vectors \(\ubold_j\), we select the
one that minimizes \(J_i^\perp(j)\), denote it by \(\vbold_i\), and update
\(\Vbold_i=[\Vbold_{i-1},\vbold_i]\).
For the resulting basis \(\Vbold=[\vbold_1,\ldots,\vbold_n]\), we form
\[
    \Shat
    =
    [\Phi(\sbold^{(1)}),\ldots,\Phi(\sbold^{(M)})]
    =
    \Vbold^\top\Sbold,
    \qquad
    \Ebold
    =
    \Sbold-\Vbold\Shat,
\]
construct an orthonormal basis and the corresponding orthogonal nonlinear feature matrix
\(\Qbold\), and obtain the nonlinear coefficient matrix directly as
\begin{equation}\label{eq:orthogonal_lifting}
    \Bboldbar
    =
    \frac{1}{M}\Ebold\Qbold^\top.
\end{equation}

\subsection{Selection of nonlinear features}\label{sec:opm:feature_selection}
An advantage of the orthogonal representation is that
the contribution of each nonlinear feature to the reconstruction error
can be quantified independently. This allows us to quantify the importance of each feature \(q_1, \dots, q_{m_n}\) of \(\qbold\) and retain only those whose inclusion yields a sufficiently large reduction in the error.

Recall that, for a fixed encoder basis \(\Vbold\), the optimal
nonlinear coefficient matrix \(\Bboldbar = [\bboldbar_1, \dots, \bboldbar_{m_n}]\) is given by~\eqref{eq:orthogonal_lifting}.
Since
\[
    \Bboldbar\qbold(\shat)
    =
    \sum_{k=1}^{m_n}
    \bboldbar_k q_k(\shat),
\]
the vector \(\bboldbar_k\) contains the coefficients associated with
the nonlinear feature \(q_k\).
Using \(\Qbold\Qbold^\top=M\Id\) and the optimality of
\(\Bboldbar\), the corresponding reconstruction objective is
\begin{equation}\label{eq:feature_full_objective}
    \frac12
    \left\|
        \Ebold-\Bboldbar\Qbold
    \right\|_\frob^2
    =
    \frac12\|\Ebold\|_\frob^2
    -
    \frac{M}{2}
    \|\Bboldbar\|_\frob^2.
\end{equation}
Since the squared Frobenius norm is the sum of the squared Euclidean
norms of the columns,
\[
    \frac{M}{2}\|\Bboldbar\|_\frob^2
    =
    \sum_{k=1}^{m_n} a_k(\Bboldbar),
    \qquad
    a_k(\Bboldbar)
    =
    \frac{M}{2}\|\bboldbar_k\|_2^2\,,
\]
we can write the reconstruction error as 
\begin{equation}\label{eq:feature_importance}
    \frac12
    \left\|
        \Ebold-\Bboldbar\Qbold
    \right\|_\frob^2
    =
    \frac12\|\Ebold\|_\frob^2
    -
    \sum_{k=1}^{m_n}a_k(\Bboldbar).
\end{equation}
Thus, \(a_k(\Bboldbar)\) is the reduction in the reconstruction objective associated
with the nonlinear feature \(q_k\), and we therefore propose to use it as a measure of the
importance of that feature.

\subsection{Greedy mode selection with inner feature selection}\label{sec:opm:greedy_feature_selection}
We now develop a feature-selection rule based on 
\Cref{sec:opm:feature_selection} and apply it during the greedy
selection of the encoder basis. For each candidate \(\ubold_j\), let
\(\Ebold^{+j}\), \(\Qbold^{+j}\), and \(\Bboldbar^{+j}\) denote the
corresponding candidate quantities, with
\[
    \Bboldbar^{+j}
    =
    \begin{bmatrix}
        \bboldbar_1^{+j}
        &
        \cdots
        &
        \bboldbar_{m_i}^{+j}
    \end{bmatrix},
    \qquad
    m_i=\binom{i+p}{p}-(i+1).
\]
We define an 
active set of features as 
\begin{equation}\label{eq:candidate_feature_importance}
    \Acal^{+j}
    =
    \left\{
        k \in \{ 1,\ldots,m_i \}
        \,|\, a_k(\Bboldbar^{+j})>\gamma
    \right\}\,,
\end{equation}
for a threshold \(\gamma > 0\). 
Let now \(\Qbold_{\Acal^{+j}}^{+j}\) contain the corresponding rows of \(\Qbold^{+j}\), and
let \(\Bboldbar_{\Acal^{+j}}^{+j}\) contain the corresponding columns of
\(\Bboldbar^{+j}\). Since every subset of rows of \(\Qbold^{+j}\) remains mutually
orthogonal, the optimal coefficient matrix for the restricted feature set is 
\begin{equation}\label{eq:restricted_lifting}
    \Bboldbar_{\Acal^{+j}}^{+j}
    =
    \frac{1}{M}
    \Ebold^{+j}(\Qbold_{\Acal^{+j}}^{+j})^\top,
\end{equation}
which is the submatrix of \(\Bboldbar^{+j}\) formed by the columns
indexed by \(\Acal^{+j}\). 
Consequently, due to the orthogonal representation of the feature map, the contribution \(a_k(\Bboldbar^{+j})\) of a feature does not depend on
which other nonlinear features are retained.

Building on the objective~\eqref{eq:greedy_objective_perp_closed}, the
feature-selected reconstruction objective for candidate \(\ubold_j\) is
\begin{equation}\label{eq:greedy_objective_feature_selection_closed}
    J_{i,\mathcal{A}^{+j}}^\perp(j)
    =
    \frac12
    \|\Ebold^{+j}\|_\frob^2
    -
    \sum_{k\in\Acal^{+j}}
    a_k(\Bboldbar^{+j}).
\end{equation}
We then select
\begin{equation}\label{eq:greedy_feature_candidate_selection}
    j_i
    \in
    \argmin_j
    J_{i,\mathcal{A}^{+j}}^\perp(j),
\end{equation}
and update
\[
    \vbold_i=\ubold_{j_i},
    \qquad
    \Vbold_i=[\Vbold_{i-1},\vbold_i].
\]

After the final encoder basis has been selected, we recompute
\(\Bboldbar=[\bboldbar_1,\ldots,\bboldbar_{m_n}]\) from~\eqref{eq:orthogonal_lifting} 
using the reduced coordinates and
orthogonal nonlinear features induced by the final basis \(\Vbold\).
We then recompute the feature importance values \(a_k(\Bboldbar)\)
and retain the corresponding active feature set analogous to~\eqref{eq:candidate_feature_importance} for threshold \(\gamma\).

\section{Lipschitz bounded orthogonal polynomial manifolds}\label{sec:lipschitz}
We now propose Greedy-OPM, which augments the orthogonal greedy construction with a sensitivity constraint on the nonlinear features to obtain polynomial decoders with an explicit bound on their Lipschitz constant.

\subsection{Lipschitz constant of the decoder}\label{sec:lipschitz_constant}
Consider the Jacobian
\[
    \Jbold_{\Gamma}(\shat)
    =
    \Vbold
    +
    \sum_{k=1}^{m_n}
    \bboldbar_k\nabla q_k(\shat)^\top
\]
of the decoder \(\Gamma\) defined in \eqref{eq:GammaBRep}. 
On a convex domain \(\Omega \subset \mathbb{R}^n\), we obtain the bound
\begin{equation}\label{eq:lipschitz_bound_generic}
    L_{\Gamma}^2
    =
    \sup_{\shat\in\Omega}
    \left\|
        \Jbold_{\Gamma}(\shat)
    \right\|_2^2
    \leq
    1
    +
    \frac{2}{M}
    \left(
        \sum_{k=1}^{m_n}
        \eta_k\sqrt{a_k(\Bboldbar)}
    \right)^2,
\end{equation}
where
\[
    \eta_k
    =
    \sup_{\shat\in\Omega}
    \|\nabla q_k(\shat)\|_2,
    \qquad
    k=1,\dots,m_n.
\]
The bound~\eqref{eq:lipschitz_bound_generic} uses that the columns of \(\Bboldbar=\frac{1}{M}\Ebold\Qbold^\top\) lie in the range of \(\Ebold=(\Id-\Vbold\Vbold^\top)\Sbold\) and are therefore orthogonal to the columns of \(\Vbold\). Hence, for every \(\xbold\in\R^n\), \(\|\Jbold_\Gamma(\shat)\xbold\|_2^2=\|\Vbold\xbold\|_2^2+\|\sum_{k=1}^{m_n}\bboldbar_k\nabla q_k(\shat)^\top\xbold\|_2^2\), where the first term equals \(\|\xbold\|_2^2\) and the second is bounded by \((\sum_{k=1}^{m_n}\|\bboldbar_k\|_2\,\eta_k)^2\|\xbold\|_2^2\) with \(\|\bboldbar_k\|_2=\sqrt{2a_k(\Bboldbar)/M}\).
Thus, the contribution of a nonlinear feature to the Lipschitz bound is
determined jointly by the magnitude of its coefficient vector \(a_k(\Bboldbar)\) and the
sensitivity \(\eta_k\) of the corresponding feature.

\subsection{Greedy-OPM: Greedy mode and feature selection under a Lipschitz budget}
\label{sec:opm:greedy_lipschitz}
We now incorporate the Lipschitz bound
\eqref{eq:lipschitz_bound_generic} directly into the greedy mode
selection.

We prescribe a Lipschitz budget \(P\geq0\). At greedy step \(i\), each previously unselected candidate \(\ubold_j\)
induces the projection error \(\Ebold^{+j}\), the orthogonal nonlinear feature
matrix \(\Qbold^{+j}\), and the sensitivities \(\eta_k^{+j}\).
For each candidate \(\ubold_j\), we define the sensitivity-constrained
objective 
\begin{equation}\label{eq:greedy_sensitivity_full}
    J_{i,P}^\perp(j)
    =
    \min_{\Bbold\in\R^{N\times m_i}}
    \frac12
    \left\|
        \Ebold^{+j}
        -
        \Bbold\Qbold^{+j}
    \right\|_\frob^2
    \quad
    \text{s.t.}
    \quad
    \sqrt{\frac{2}{M}}
\sum_{k=1}^{m_i}
\eta_k^{+j}
\sqrt{a_k(\Bbold)}
\leq P.
\end{equation}
The next encoder mode is then selected as
\[
    j_i
    \in
    \argmin_j
    J_{i,P}^\perp(j).
\]
Thus, for every candidate encoder direction, the inner problem determines
the nonlinear coefficients that give the smallest error
while satisfying the prescribed Lipschitz budget \(P\).

The inner optimization problem in
\eqref{eq:greedy_sensitivity_full} can be simplified using the orthogonal
representation. The objective can be written as
\begin{equation}\label{eq:constrained_square_completion}
    \frac12
    \left\|
        \Ebold^{+j}
        -
        \Bbold\Qbold^{+j}
    \right\|_\frob^2
    =
    \frac12
    \left\|
        \Ebold^{+j}
        -
        \Bboldbar^{+j}\Qbold^{+j}
    \right\|_\frob^2
    +
    \frac{M}{2}
    \left\|
        \Bbold-\Bboldbar^{+j}
    \right\|_\frob^2,
\end{equation}
where
\(\Bboldbar^{+j}
=
\frac{1}{M}
\Ebold^{+j}
(\Qbold^{+j})^\top\)
denotes the unconstrained least-squares solution.
Since the first term in
\eqref{eq:constrained_square_completion} is independent of
\(\Bbold\), the inner problem in
\eqref{eq:greedy_sensitivity_full} has the same minimizer as
\begin{equation}\label{eq:greedy_sensitivity_inner}
    \Bboldbar_P^{+j} = \argmin_{\Bbold}
    \frac12
    \left\|
        \Bbold-\Bboldbar^{+j}
    \right\|_\frob^2
    \qquad
    \text{s.t.}
    \qquad
    \sqrt{\frac{2}{M}}
\sum_{k=1}^{m_i}
\eta_k^{+j}
\sqrt{a_k(\Bbold)}
\leq P.
\end{equation}
Since the sensitivity constraint in
\eqref{eq:greedy_sensitivity_inner} is equivalent to
\[
    \sum_{k=1}^{m_i}
    \eta_k^{+j}\|\bbold_k\|_2
    \leq P,
\]
we introduce the weighted mixed norm
\[
    \|\Bbold\|_{2,1,\boldsymbol{\eta}^{+j}}
    =
    \sum_{k=1}^{m_i}
    \eta_k^{+j}\|\bbold_k\|_2.
\]
Hence, the feasible set is the weighted group-\(\ell_{2,1}\) ball
\[
    \mathcal{B}_P^{+j}
    =
    \left\{
        \Bbold\in\R^{N\times m_i}
        \,|\,
        \|\Bbold\|_{2,1,\boldsymbol{\eta}^{+j}}
        \leq P
    \right\},
\]
and \eqref{eq:greedy_sensitivity_inner} is the Euclidean projection of
\(\Bboldbar^{+j}\) onto \(\mathcal{B}_P^{+j}\).
This projection is characterized by a single Lagrange multiplier
\(\lambda\geq0\). For fixed \(\lambda\), the problem separates over the
coefficient columns and the solution is given by group soft-thresholding,
\begin{equation}\label{eq:soft-threshold}
    \bboldbar_{P,k}^{+j}
    =
    \left(
        1-
        \frac{\lambda\eta_k^{+j}}
             {\|\bboldbar_k^{+j}\|_2}
    \right)_+
    \bboldbar_k^{+j},
    \qquad
    k=1,\ldots,m_i\,,
\end{equation}
where \((x)_+=\max\{x,0\}\) denotes the positive-part operator~\cite{parikh2014proximal}. 
If the unconstrained solution is feasible, then \(\lambda=0\). Otherwise,
\(\lambda>0\) is chosen so that the sensitivity constraint is active. The
corresponding scalar equation is continuous, monotone, and piecewise
linear in \(\lambda\), with breakpoints  \(\|\bboldbar_k^{+j}\|_2/\eta_k^{+j}.\) 
We compute \(\lambda\) using the sorted-breakpoint projection procedure
of~\cite{duchi2008Efficient,condat2016Fast}, after which
\(\Bboldbar_P^{+j}\) is obtained in the single group soft-thresholding step~\eqref{eq:soft-threshold}. Thus, no iterative optimization over the full
coefficient matrix is required.
The corresponding objective is
\begin{equation}\label{eq:greedy_sensitivity_objective}
    J_{i,P}^\perp(j)
    =
    \frac12
    \left\|
        \Ebold^{+j}
        -
        \Bboldbar_P^{+j}\Qbold^{+j}
    \right\|_\frob^2\,.
\end{equation}
This objective is minimized over the candidate directions at each greedy step.

During the greedy iterations, no
explicit active set is required, since these zero columns do not contribute
to the candidate score. 
Nevertheless, the group soft-thresholding~\eqref{eq:soft-threshold}
implies, for each candidate \(\ubold_j\), the active feature set
\begin{equation}\label{eq:lipschitz_active_set_candidate}
    \Acal_P^{+j}
    =
    \left\{
        k \in \{ 1,\ldots,m_i \}
        \,|\, a_k(\Bboldbar^{+j}) > \gamma_k^{+j}
    \right\},
    \qquad
    \gamma_k^{+j}
    =
    \frac{M\,(\lambda^{+j}\eta_k^{+j})^2}{2},
\end{equation}
where \(\lambda^{+j}\geq0\) is the Lagrange multiplier of~\eqref{eq:greedy_sensitivity_inner} for candidate \(\ubold_j\), and \(\bboldbar_{P,k}^{+j}=\0\) exactly when
\(k\notin\Acal_P^{+j}\). This establishes a clear connection to~\eqref{eq:candidate_feature_importance},
with the single prescribed threshold \(\gamma\) replaced by thresholds
\(\gamma_k^{+j}\) that are determined by the budget \(P\) through
\(\lambda^{+j}\) and grow with the sensitivity \(\eta_k^{+j}\) of the feature. In essence, the control of the Lipschitz constant is tantamount to the control of the threshold, which in turn leads to the selection of features to satisfy the prescribed budget \( P \).

After the final encoder basis vector has been selected, we
compute the sensitivity-constrained coefficient matrix
\(\Bboldbar_P=[\bboldbar_{P,1},\ldots,\bboldbar_{P,m_n}]\) as in~\eqref{eq:greedy_sensitivity_inner} from \(\Bboldbar\) and define the
active feature set as
\[
    \Acal_P
    =
    \left\{
        k\in\{1,\ldots,m_n\}
        \,|\,
        \|\bboldbar_{P,k}\|_2>0
    \right\}
\]
or, equivalently, as in~\eqref{eq:lipschitz_active_set_candidate},
\[
    \Acal_P
    =
    \left\{
        k\in\{1,\ldots,m_n\}
        \,|\,
        a_k(\Bboldbar) > \gamma_k
    \right\},
    \qquad
    \gamma_k = \frac{M\lambda^2\eta_k^2}{2},
\]
where \(\lambda\geq0\) is the multiplier of the final projection of
\(\Bboldbar\) onto \(\mathcal{B}_P\).
The final decoder then retains only the nonlinear features indexed by
\(\Acal_P\). Importantly, the final constrained coefficient matrix satisfies
\[
    \sqrt{\frac{2}{M}}
    \sum_{k=1}^{m_n}
    \eta_k
    \sqrt{a_k(\Bboldbar_P)}
    \leq P.
\]
Together with~\eqref{eq:lipschitz_bound_generic}, which applies because the group soft-thresholding only rescales the columns of \(\Bboldbar\) and thus preserves their orthogonality to the columns of \(\Vbold\), this guarantees a bound on the Lipschitz constant of the decoder, 
\[
    L_{\Gamma}^2
    \leq
    1+P^2
\]
on the domain \(\Omega\).

\subsection{Implementation}
\label{sec:lipschitz:algorithm} We summarize three implementation details that improve the efficiency of the
construction. An implementation is available at
\url{https://github.com/smallpondtom/greedyopm}.

First, direct evaluation of \(J_{i,P}^\perp(j)\) would require forming the
high-dimensional residual \(\Ebold^{+j}\in\R^{N\times M}\) for every
candidate. We avoid this by using the rank-\(s\) truncated singular value
decomposition
\(\Ubold\Sigmasmat\Wbold^\top\) of \(\Sbold\), whose left singular vectors
provide the candidate encoder modes.
Let \(\Jcal_{i-1}\subset\{1,\ldots,s\}\) denote the indices selected
after \(i-1\) greedy steps, and define
\(\Jcal^{+j}=\Jcal_{i-1}\cup\{j\}\) and
\(\Jcal^{-j}=\{1,\ldots,s\}\setminus\Jcal^{+j}\).
Using the orthonormality of the columns of \(\Ubold\), the candidate reduced
coordinates and residual norm are obtained directly from the SVD as
\[
    \Shat^{+j}
    =
    \Sigmasmat_{\Jcal^{+j}}
    \Wbold_{\Jcal^{+j}}^\top,
    \qquad
    \|\Ebold^{+j}\|_\frob^2
    =
    \sum_{\ell\in\Jcal^{-j}}\sigma_\ell^2.
\]
The truncation to rank \(s\) drops the singular directions \(\ell>s\)
from every candidate residual, which avoids the full SVD of \(\Sbold\).
Since every candidate is scored under the same truncation, the comparison
is fair for any \(s\): the dropped directions add the same constant
\(\sum_{\ell>s}\sigma_\ell^2\) to every \(\|\Ebold^{+j}\|_\frob^2\), and
they perturb the feature importance values below by at most
\(\sigma_{s+1}^2/2 \), which is negligible
once \(\sigma_{s+1}\) is small relative to the retained singular values. 
Increasing \( s \) tightens this bound and thus improves the accuracy of
the candidate scores, at the cost of a larger candidate pool to evaluate
at every greedy step.

Second, the feature importance values can be evaluated without forming
\(\Ebold^{+j}\),
\begin{equation}\label{eq:svd_feature_importance}
    a_k(\Bboldbar^{+j})
    =
    \frac{1}{2M}
    \left\|
        \Sigmasmat_{\Jcal^{-j}}
        \Wbold_{\Jcal^{-j}}^\top
        [\Qbold^{+j}]_{k,:}^\top
    \right\|_2^2,
    \qquad
    k=1,\ldots,m_i.
\end{equation}
The feature matrix \(\Qbold^{+j}\) and the sensitivities
\(\eta_k^{+j}\) are computed entirely in the reduced coordinates.
Furthermore, the group soft-thresholding step depends on
\(\Bboldbar^{+j}\) only through its column norms,
\[
    \|\bboldbar_k^{+j}\|_2
    =
    \sqrt{\frac{2}{M}a_k(\Bboldbar^{+j})}.
\]
Hence, the sensitivity-constrained candidate score
\(J_{i,P}^\perp(j)\) can be evaluated from
\(\|\Ebold^{+j}\|_\frob^2\),
\(\{a_k(\Bboldbar^{+j})\}\),
\(\{\eta_k^{+j}\}\), and \(P\), without forming any
\(N\)-dimensional candidate coefficient matrix.

Third, the truncated SVD is used only to evaluate and compare candidate
encoder modes. After all \(n\) modes have been selected, the final residual,
orthogonal features, and coefficient matrix are recomputed from the full
data. The sensitivity constraint is then applied once more to obtain the
final coefficient matrix \(\Bboldbar_P\) and active feature set
\(\Acal_P\).

%% file: sections/numerics.tex
\section{Numerical experiments}\label{sec:numerics}
We now demonstrate Greedy-OPM with three numerical experiments: a one-dimensional linear advection-diffusion equation, the two-dimensional viscous Burgers equations, and a two-dimensional Boussinesq rising thermal plume.

\subsection{General setup}\label{sec:numerics:setup}
The general setup is as follows. 
\subsubsection{Compared methods}
We compare Greedy-OPM to the baseline linear encoder-decoder pair (POD), the approach of~\cite{geelen2023Operator}, which uses the POD basis vectors as the encoder modes of a quadratic manifold whose coefficient matrix with group-sparse penalization is solved by \( 400 \) proximal-gradient iterations (POD-Encoder QM), and the plain greedy method of~\cite{schwerdtner2026Greedy} with the full monomial feature set of maximal degree \( p=2 \) and \( p=3 \) (Plain Greedy \( (p=2) \) and \( (p=3) \)). Greedy-OPM is likewise run with maximal degree \( p=2 \) and \( p=3 \). All of these models are fitted with Tikhonov regression using a regularization parameter of \( 10^{-13} \).

We further compare to the sparse compositional polynomial network of~\cite{bensalah2026Nonlinear} (CPN-S), using the implementation released by the authors\footnote{\url{https://github.com/JoelSOFFO/CPN_MOR}}. 
We report \( p=3 \) in the main text and \( p=5 \) in \Cref{app:extra_plots}.  
We also stop the selection whenever \( n \) reaches \( 10 \) to match the other methods.
Because the component maps are composed, the decoder of CPN-S typically is a polynomial of degree higher than $p$ in the reduced coordinates, and each trailing coefficient has its own set of basis functions, so its feature count is not comparable to the shared feature sets of the other methods and is reported in the text instead of the feature plots.

\subsubsection{Data setup}
The problems that we consider depend on a scalar physics parameter \( \mu \). The training data are obtained at four values of \( \mu \) spaced equidistantly in the training parameter domain, equidistantly in \( \log\mu \) for the advection-diffusion example. The validation data consist of three parameter values different from those of the training. Two of the parameters are selected within the training parameter range and one parameter value is selected outside the training range. For testing, we use one new parameter value within the training range and one value outside.

For the training-testing data split with snapshots \( \sbold^{(1)},\ldots,\sbold^{(M)} \) and shifted data matrix \( \Sbold \) defined with respect to the training mean \( \sref \), we report the relative Frobenius reconstruction error
\begin{equation}\label{eq:rel_err}
    \Ecal(\Sbold)
    =
    \frac{
        \left(
            \sum_{\ell=1}^{M}
            \left\|
                \sbold^{(\ell)}
                -
                \Gamma\!\left(\Phi(\sbold^{(\ell)})\right)
            \right\|_2^2
        \right)^{1/2}
    }{
        \|\Sbold\|_\frob
    }.
\end{equation}
Thus, the numerator measures the reconstruction error of the complete encoder-decoder map \( \Gamma\circ\Phi \), while the denominator measures the variation of the data about the training reference state.

All greedy methods use a candidate pool of size \( s=50 \), which means that we compute a rank-\( s \) truncated singular value decomposition of the shifted training data matrix and use its first \( s \) left singular vectors as candidate encoder modes \( \ubold_1, \dots, \ubold_{s} \), and the reduced dimension is swept over \( n=1,\ldots,10 \).

\subsubsection{Lipschitz bound settings}
The quantities \( \eta_k \) entering the Lipschitz bound require the maximum of \( \|\nabla q_k(\shat)\|_2 \) over a prescribed reduced-coordinate domain. Let \( \Omega_0 \) denote the smallest axis-aligned box containing the reduced training coordinates and let \( \shat_c \) denote its center. For a dilation factor \( \delta\geq 0 \), we define
\[
    \Omega_\delta
    =
    \shat_c
    +
    (1+\delta)(\Omega_0-\shat_c).
\]
We approximate each \( \eta_k \) by the largest value of \( \|\nabla q_k(\shat)\|_2 \) over the dilated reduced training coordinates \( \shat_c+(1+\delta)(\shat^{(\ell)}-\shat_c)\in\Omega_\delta \) of an equispaced subsample of \( 250 \) training snapshots. Throughout all experiments we use \( \delta=0.1 \), so that the domain extends \( 10\% \) beyond the reduced training domain in every direction from its center. During the greedy iterations, this construction is applied to the reduced coordinates associated with each candidate basis.

We select the Lipschitz budget \(P\) using the validation data. This plays the
same role as tuning the regularization parameters of the competing manifold
methods, such as the group-sparsity weight in POD-Encoder QM or the tolerances
in the CPN-S algorithm. The value of $P$ is selected separately for the maximal polynomial degree \( p=2 \) and \( p=3 \) of the feature map. 
We select a single budget that accounts for both interpolation and extrapolation. For each trial value of \( P \), we compute the reconstruction error \( \Ecal_{\mathrm{in}}(P) \) on the two interior validation parameter values and the reconstruction error \( \Ecal_{\mathrm{ex}}(P) \) on the validation parameter value beyond the training range, and minimize
\[
    \Ecal_{\mathrm{val}}(P)
    =
    \sqrt{
        \Ecal_{\mathrm{in}}(P)
        \Ecal_{\mathrm{ex}}(P)
    }.
\]
The geometric mean gives equal weight to the two regimes on a logarithmic scale,
\[
    \log \Ecal_{\mathrm{val}}
    =
    \frac{1}{2}
    \left(
        \log \Ecal_{\mathrm{in}}
        +
        \log \Ecal_{\mathrm{ex}}
    \right),
\]
and therefore avoids selecting the budget solely according to the regime with the larger absolute error. The budget is selected by a grid-adaptive sweep over the Lipschitz bound \( \sqrt{1+P^2} \) of~\Cref{sec:opm:greedy_lipschitz}, which we vary over a logarithmic grid on \( [1.05,10^5] \) (corresponding to \( P\in[0.32,10^5] \)), bracketing the minimizer by its neighboring grid points, and refining on linear grids inside this bracket until the grid spacing reaches \( 10^{-2} \). Each trial budget recalibrates the existing fit through~\eqref{eq:greedy_sensitivity_inner} without re-selecting the encoder modes, so the sweep takes less than one minute in all examples. The budget is selected once at \( n=10 \) and the resulting value is then kept fixed for the complete sweep \( n=1,\ldots,10 \), so that changes with \( n \) reflect the reduced dimension rather than repeated hyperparameter retuning. For each example we show the validation objective \( \Ecal_{\mathrm{val}}(P) \) and the number of active nonlinear features over the budget sweep at \( n=10 \).

\subsection{Linear advection-diffusion}\label{sec:numerics:lad}
The first example we consider is the one-dimensional linear advection-diffusion equation. 

\subsubsection{Linear advection-diffusion: Setup}

The periodic linear advection-diffusion equation is
\begin{equation}\label{eq:lad}
    u_t+c\,u_x = \nu\,u_{xx}, \qquad x\in[0,1), \qquad c=1, \qquad \nu=10^{-4},
\end{equation}
with a Gaussian pulse \( u_0(x)=(\pi w)^{-1/2}\exp(-(x-x_0)^2/w) \) of squared width \( w \) centered at \( x_0=0.2 \). Each Fourier mode evolves exactly, \( \hat u_k(t)=\hat u_k(0)\exp(-(2\pi\mathrm{i}kc+\nu(2\pi k)^2)t) \), so the snapshots are generated without time-integration error on \( N=1024 \) grid points over \( t\in[0,1) \), one traversal of the domain.
The varied parameter is the pulse width \( w \). A narrower pulse carries more Fourier modes of significant amplitude, so a fixed number of \gls{pod} modes captures its transport to a lesser extent, and the extrapolation is directed toward a pulse narrower than any pulse in the training data. Training uses \( w\in\{1,\,1.59,\,2.52,\,4\}\times10^{-3} \) with \( 120 \) snapshots each (\( M=480 \)), validation uses \( w\in\{1.26,\,3.17\}\times10^{-3} \) and \( w=0.8\times10^{-3} \) with \( 100 \) snapshots each, and the tests use \( w=2.81\times10^{-3} \) (within the training range) and \( w=0.5\times10^{-3} \) (outside the training range) with \( 300 \) snapshots each. Budget selection returns \( P^\star=190.89 \) for \( p=2 \) and \( P^\star=1798.95 \) for \( p=3 \).

\subsubsection{Linear advection-diffusion: Results}

\begin{figure}[tb!]
    \centering
    \includegraphics[width=\textwidth]{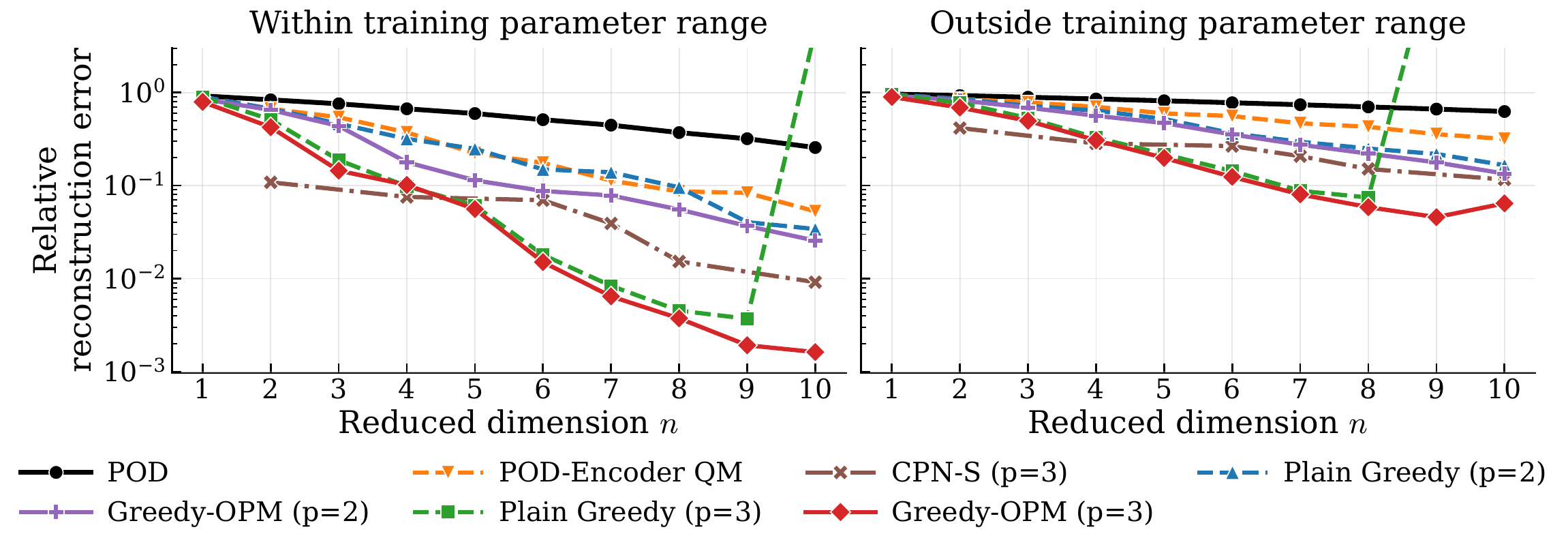}
    \caption{Linear advection-diffusion. Relative reconstruction error~\eqref{eq:rel_err} against the reduced dimension \( n \) on the test parameters within (\( w=2.81\times10^{-3} \)) and outside (\( w=0.5\times10^{-3} \)) the training range.}\label{fig:lad_error}
\end{figure}

\begin{figure}[tb]
    \centering
    \includegraphics[width=\textwidth]{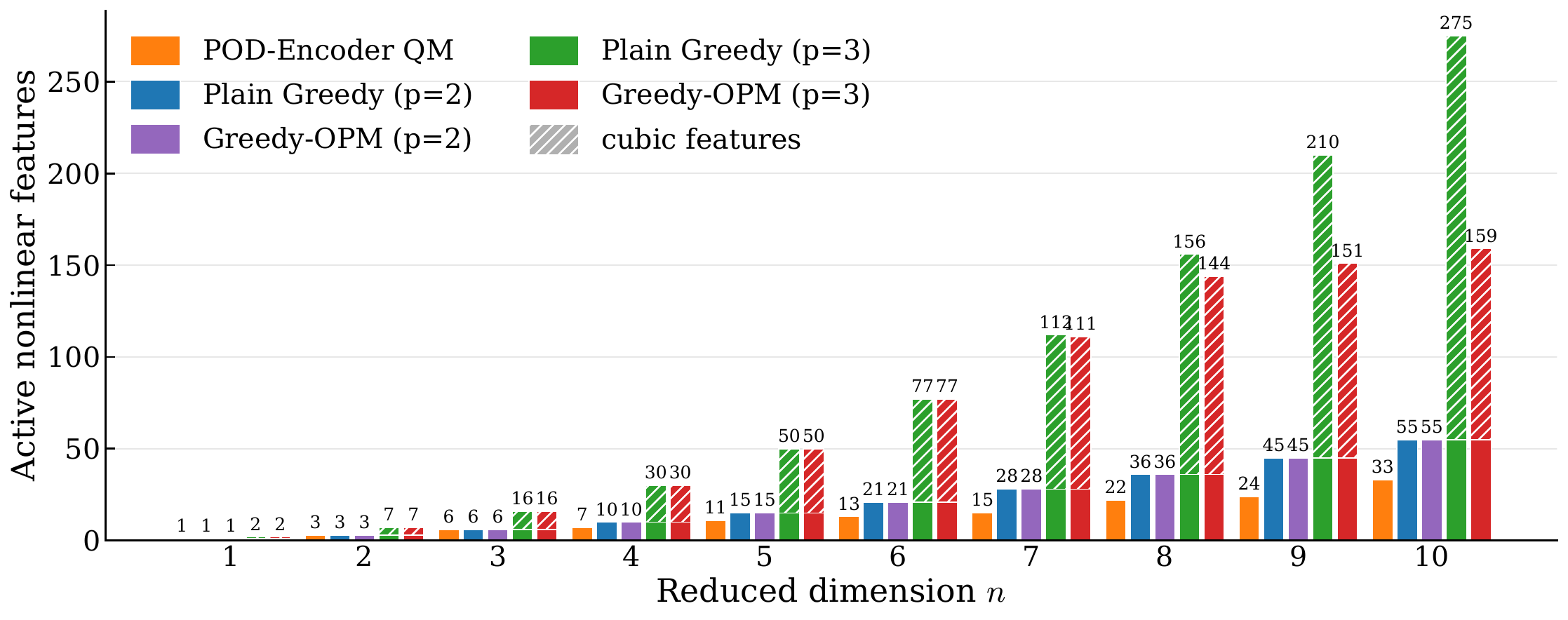}
    \caption{Linear advection-diffusion. Active nonlinear features against \( n \), with the Greedy-OPM \( (p=3) \) bars split into degree-2 and degree-3 contributions.}\label{fig:lad_active}
\end{figure}

\begin{figure}[tb]
    \centering
    \includegraphics[width=\textwidth]{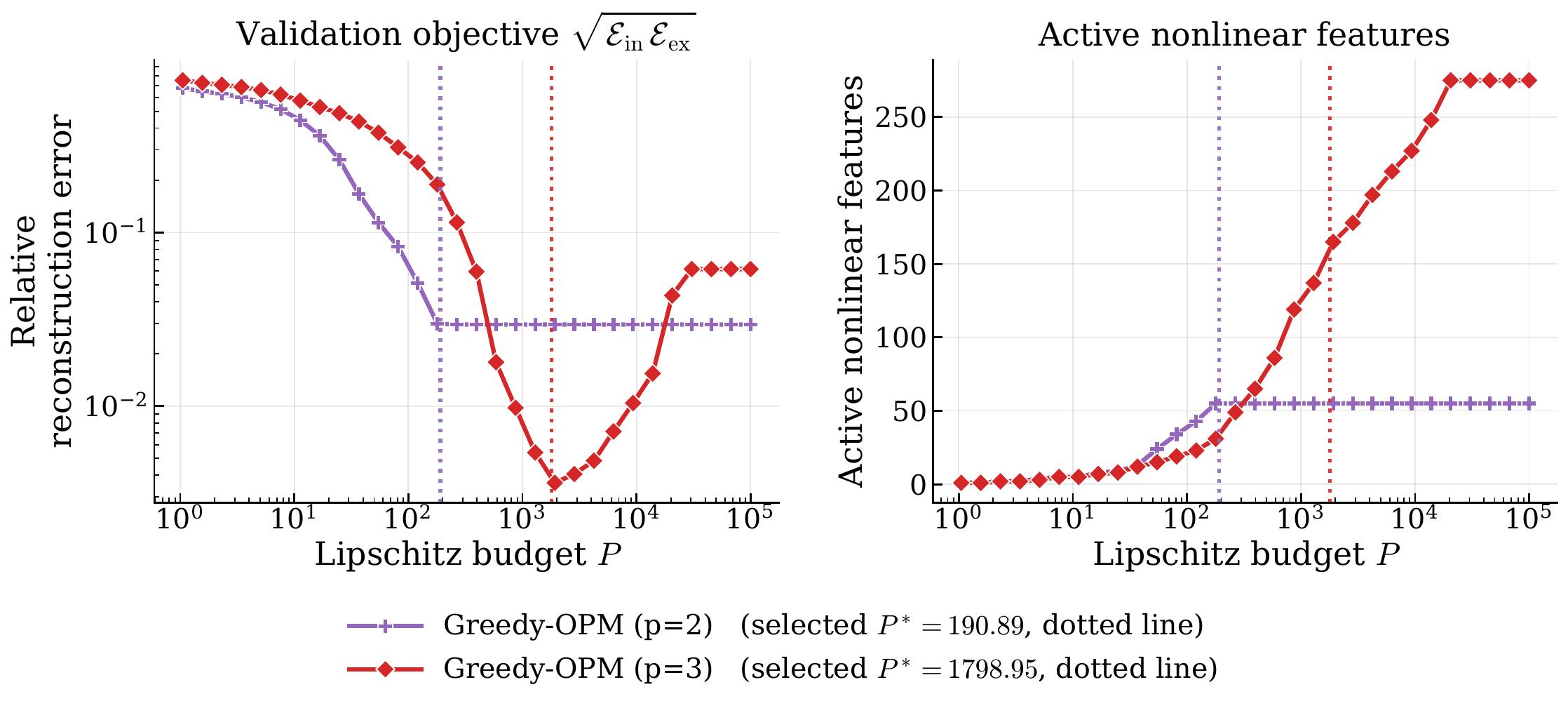}
    \caption{Linear advection-diffusion. Validation objective \( \Ecal_{\mathrm{val}} \) and number of active nonlinear features against the Lipschitz budget \( P \) at \( n=10 \). Dotted lines mark the selected \( P^\star \) of each variant.}\label{fig:lad_budget}
\end{figure}

\begin{figure}[htbp!]
    \centering
    \includegraphics[width=\textwidth]{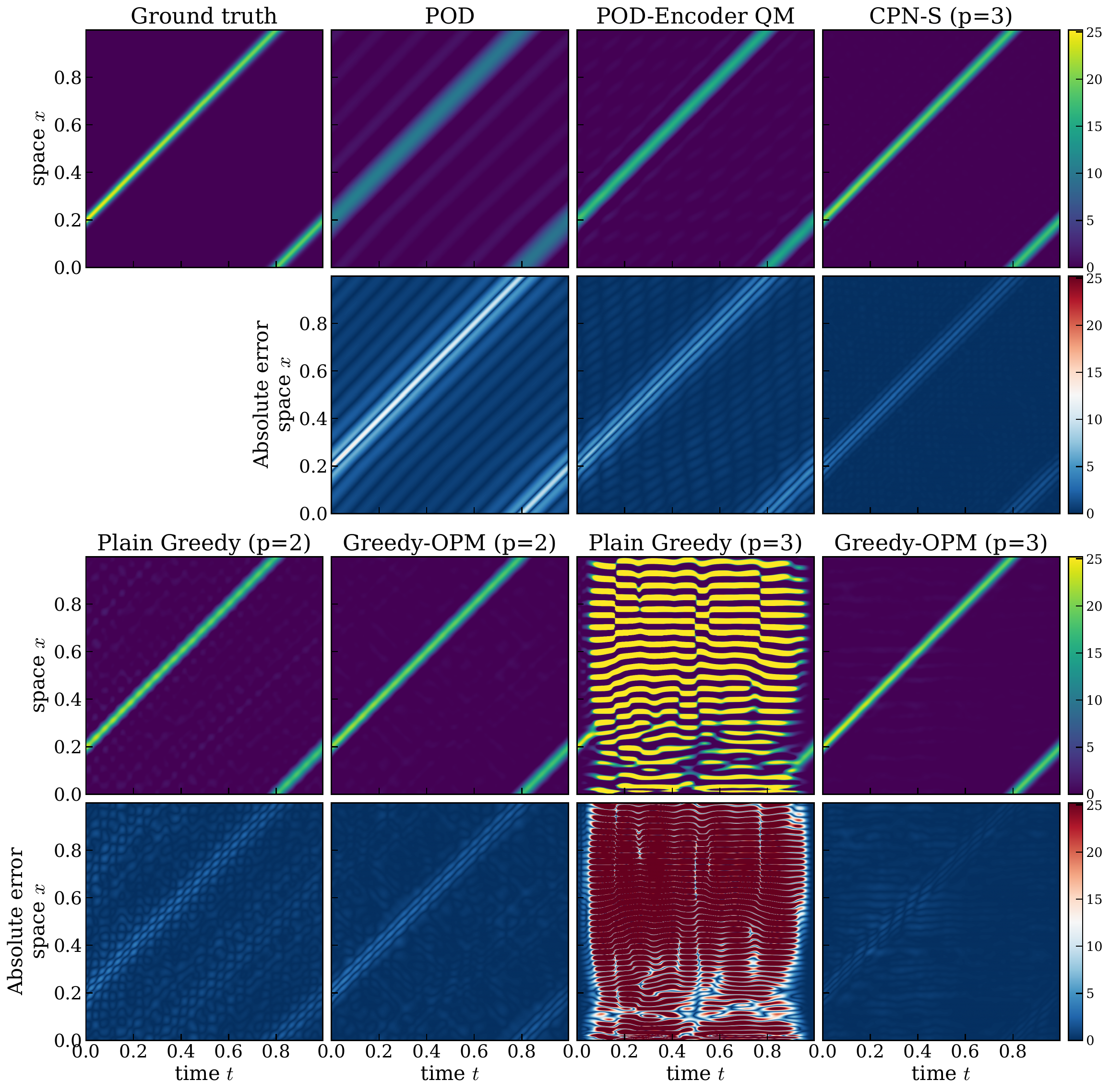}
    \caption{Linear advection-diffusion. Space-time field \( u(x,t) \) of the test parameter outside the training range (\( w=0.5\times10^{-3} \)) reconstructed at \( n=10 \), with the absolute error below each reconstruction.}\label{fig:lad-field}
\end{figure}

\Cref{fig:lad_error} shows the error against \( n \) for the two test parameters. For Plain Greedy \( (p=3) \), we observe a clear turnover in the test error from \( n = 8 \) on for both test cases, ending more than an order of magnitude above \gls{pod} at \( n=10 \), illustrating its sensitivity to deviations from the training data. In contrast, Greedy-OPM \( (p=3) \) achieves a decreasing test error for the interpolated test parameter case for all \( n \), and has the lowest error of all methods from \( n=5 \) on. 
The test errors of quadratic manifolds are larger than their cubic counterparts and are all on par with each other while remaining stable for all \( n \). The baseline POD shows the highest error.

\Cref{fig:lad_active} shows the active features over the reduced dimensions. Greedy-OPM \( (p=3) \) retains \( 159 \) of the \( 275 \) candidate features at \( n=10 \), \( 55 \) quadratic and \( 104 \) cubic, so the budget removes more than half of the \( 220 \) cubic features that Plain Greedy \( (p=3) \) uses. This pruning of features is what prevents the error turnover of Greedy-OPM \( (p=3) \). CPN-S uses \( 219 \) distinct nonlinear basis functions at \( n=10 \).

\Cref{fig:lad_budget} shows the validation objective over the budget sweep at \( n=10 \). For \( p=3 \) the objective has a distinct minimum at \( P^\star \) and rises toward the unconstrained end of the sweep, where all \( 275 \) features are active. For \( p=2 \) the objective decreases monotonically and saturates once all \( 55 \) quadratic features are active, so the quadratic manifold does not overfit this data and the budget is inactive for \( P>P^\star \).

\Cref{fig:lad-field} shows the reconstructed space-time field of the pulse outside the training range. \Gls{pod} and POD-Encoder QM smear the pulse into parallel streaks along its trajectory, and Plain Greedy \( (p=3) \) produces oscillations across the whole domain whose amplitude exceeds the pulse itself. Greedy-OPM \( (p=3) \) and CPN-S recover the sharp pulse, and their remaining error is focused on a thin band along the pulse.

\subsection{Viscous Burgers}\label{sec:numerics:burgers}
We now consider the two-dimensional Burgers equation.

\subsubsection{Viscous Burgers: Setup}
The viscous Burgers equations on the periodic square \( [0,2\pi)^2 \) are
    \begin{equation}\label{eq:burgers}
        u_t+u\,u_x+v\,u_y = \nu\,\Delta u,
        \qquad
        v_t+u\,v_x+v\,v_y = \nu\,\Delta v,
    \end{equation}
    initialized by a two-scale vortical perturbation of a uniform mean flow \( (c,c) \) with \( c=1 \),
    \begin{displaymath}
        u_0 = c+\sin x\cos y+\tfrac12\sin 2x\cos 3y,
        \qquad
        v_0 = c-\cos x\sin y+\tfrac25\cos 3x\sin 2y .
    \end{displaymath}
    The nonlinearity steepens these harmonics into thin internal layers whose thickness is set by \( \nu \), and the mean flow sweeps the pattern diagonally across the box, which is the transport mechanism behind the slow \( n \)-width decay.

The solver is Fourier pseudo-spectral with the quadratic terms formed in physical space and 2/3-dealiased~\cite{canuto2007Spectral,orszag1971Eliminationa}, and explicit RK4 in time with \( \Delta t=10^{-3} \), integrated to \( t=2\pi \), one full traversal of the box at the mean-flow speed. The grid has \( 128 \) points per direction and the state stacks both velocity components, so \( N=2\cdot128^2=32768 \).

Because a constant mean flow only translates the solution, the Reynolds number \( Re=2\pi/\nu \) rather than the advection speed is the parameter that changes the shape of the solution, and it is what the experiment varies. Training uses \( Re\in\{140,206.7,273.3,340\} \) with \( 85 \) snapshots each (\( M=340 \)), validation uses \( Re\in\{173.3,306.7\} \) and \( Re=460 \), and the tests use \( Re=264.7 \) (within the training range) and \( Re=580 \) (outside the training range) with \( 131 \) snapshots each.

The training set is deliberately sparse relative to the feature count, with \( M=340 \) columns against \( m_n=275 \) features at \( n=10 \) and \( p=3 \). This is above the threshold at which the empirical inner product~\eqref{eq:emp_ip} is positive definite, but close enough to it that a degree-3 manifold can fit the training data and fail elsewhere.

Budget selection returns \( P^\star=393.10 \) for \( p=2 \) and \( P^\star=1558.95 \) for \( p=3 \).

\subsubsection{Viscous Burgers: Results}

\begin{figure}[tb]
    \centering
    \includegraphics[width=\textwidth]{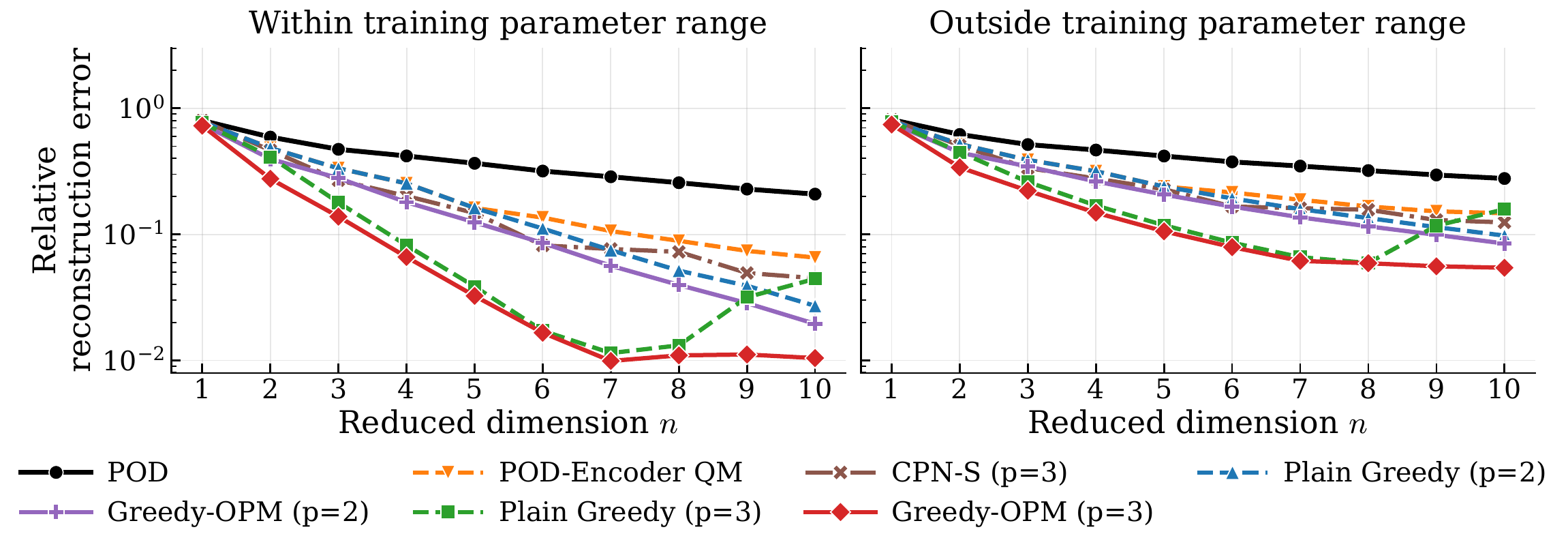}
    \caption{Two-dimensional viscous Burgers. Relative reconstruction error~\eqref{eq:rel_err} against \( n \) on the test parameters within (\( Re=264.7 \)) and outside (\( Re=580 \)) the training range.}\label{fig:bur_error}
\end{figure}

\begin{figure}[tb]
    \centering
    \includegraphics[width=\textwidth]{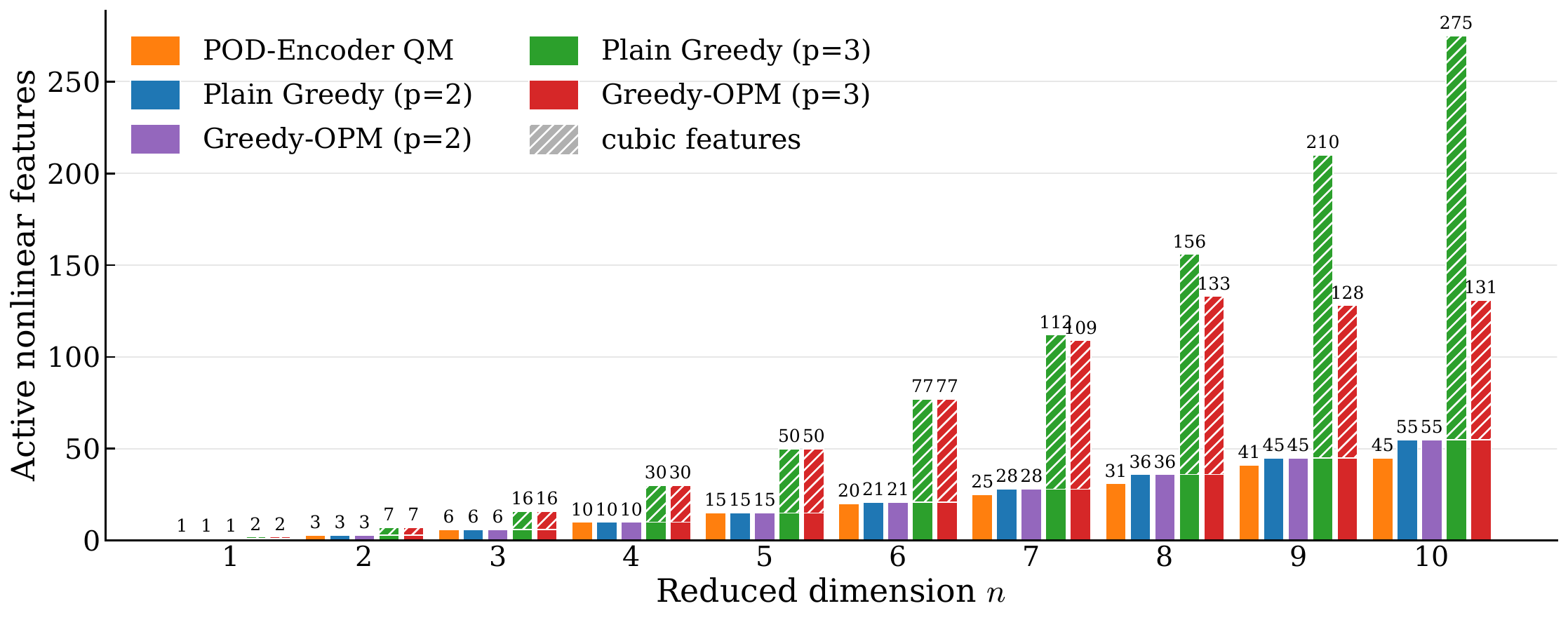}
    \caption{Two-dimensional viscous Burgers. Active nonlinear features against \( n \), degree-split for Greedy-OPM \( (p=3) \).}\label{fig:bur_active}
\end{figure}

\begin{figure}[htbp!]
    \centering
    \includegraphics[width=\textwidth]{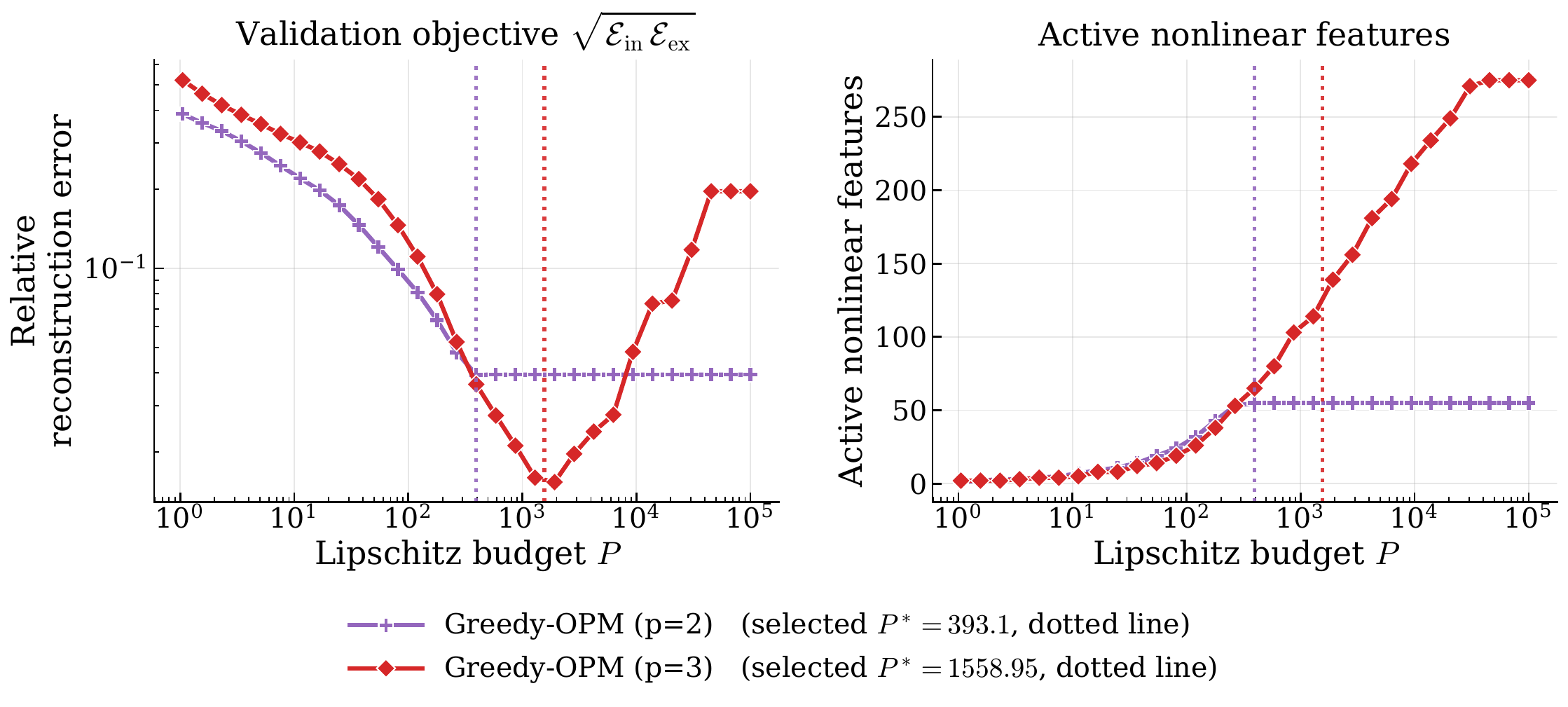}
    \caption{Two-dimensional viscous Burgers. Validation objective \( \Ecal_{\mathrm{val}} \) and number of active nonlinear features against the Lipschitz budget \( P \) at \( n=10 \), with the selected \( P^\star \) marked.}\label{fig:bur_budget}
\end{figure}

\begin{figure}[htbp!]
    \centering
    \includegraphics[width=\textwidth]{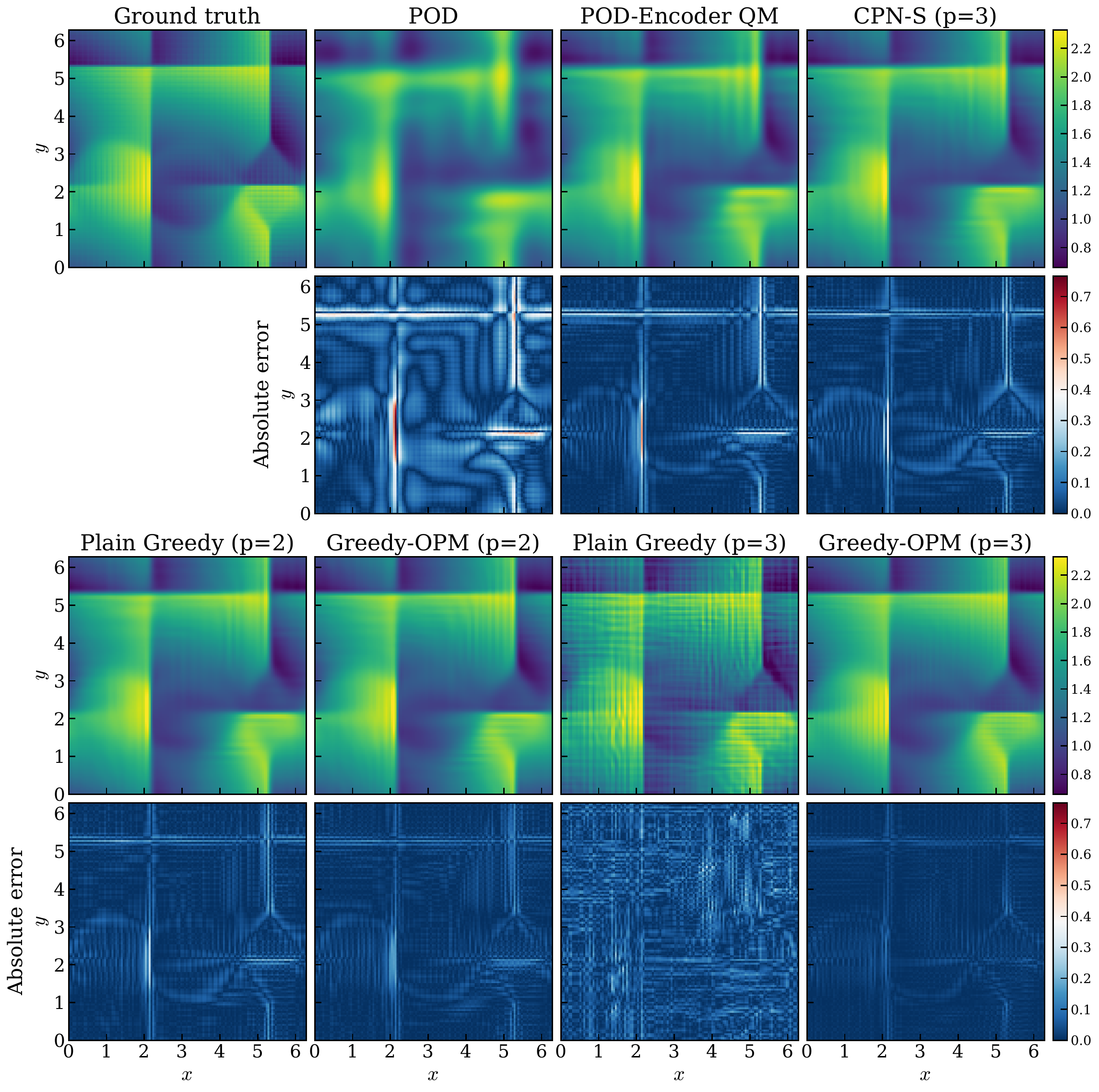}
    \caption{Two-dimensional viscous Burgers. Speed \( \sqrt{u^2+v^2} \) of the snapshot least captured by the linear \gls{pod} subspace at the test parameter outside the training range (\( Re=580 \)), reconstructed at \( n=10 \), with the absolute error below each reconstruction.}\label{fig:bur-field}
\end{figure}

\Cref{fig:bur_error} shows the error against \( n \) for the two test parameters. Greedy-OPM \( (p=3) \) has the lowest error in both test cases for all \( n \), apart from \( n=8 \) outside the training range, where Plain Greedy \( (p=3) \) is marginally lower, with Plain Greedy \( (p=3) \) on par with it until Plain Greedy leads to high errors after $n > 7$.  
The error of Greedy-OPM \( (p=3) \) levels off from \( n = 7 \) on in both test cases, and the quadratic manifolds, which keep improving, narrow the gap. The parametric variation of this flow lies in the thickness of the internal layers, which four training Reynolds numbers sample only coarsely, so the cubic features added beyond \( n = 7 \) fit the training layers rather than their dependence on \( Re \). This is the overfitting of Plain Greedy \( (p=3) \). The budget removes these features from Greedy-OPM \( (p=3) \), so its error flattens where Plain Greedy \( (p=3) \) starts to show instabilities.

\Cref{fig:bur_active} shows the active features over the reduced dimensions. Greedy-OPM \( (p=3) \) keeps \( 131 \) of the \( 275 \) candidate features at \( n=10 \), \( 55 \) quadratic and \( 76 \) cubic, so again more than half of the \( 220 \) cubic features of Plain Greedy \( (p=3) \) are removed. From \( n=8 \) on, the size of its active set stays nearly constant while the quadratic share grows, so the budget trades cubic for quadratic features once the linear part carries more of the reconstruction. CPN-S uses \( 117 \) distinct nonlinear basis functions at \( n=10 \). 

\Cref{fig:bur_budget} shows the validation objective over the budget sweep at \( n=10 \). For \( p=3 \) the objective has a distinct minimum at \( P^\star \) and rises by more than one order of magnitude toward the unconstrained end of the sweep, where all \( 275 \) features are active. 

\Cref{fig:bur-field} shows the speed field of the worst-approximated snapshot outside the training range. The error of \gls{pod} is concentrated on the internal layers, which it smears, Plain Greedy \( (p=3) \) adds grid-scale oscillations over the whole box, and Greedy-OPM \( (p=3) \) leaves the smallest error, confined to the thinnest layers.

\subsection{Boussinesq rising thermal plume}\label{sec:numerics:boussinesq}
We now consider the incompressible Boussinesq equations describing a thermal plume.
\subsubsection{Boussinesq rising thermal plume: Setup}
The Boussinesq equations in dimensionless form~\cite{turner1957Buoyant,spiegel1960Boussinesq,tritton1977Physical} are
    \begin{equation}\label{eq:boussinesq}
        \begin{aligned}
            \ubold_t+(\ubold\cdot\nabla)\ubold &= -\nabla p+\frac{1}{Re}\Delta\ubold+\theta\,\ebold_y,
            \qquad
            \nabla\cdot\ubold=0,
            \\
            \theta_t+\ubold\cdot\nabla\theta &= \frac{1}{Re\,Pr}\Delta\theta,
        \end{aligned}
    \end{equation}
    with velocity \( \ubold=(u,v) \) and temperature perturbation \( \theta \) on the periodic box \( [0,8)\times[0,14) \). Lengths are scaled by the radius of the initial thermal and velocities by the buoyancy velocity, so that the Reynolds number \( Re \) and the Prandtl number \( Pr=1 \) are the only parameters. The fluid starts at rest with a Gaussian thermal \( \theta_0=\exp(-(x-4)^2-(y-3)^2) \), which accelerates upward, rolls up into a buoyant vortex pair, and sheds secondary rolls along its cap and stem as \( Re \) grows.

The solver is Fourier pseudo-spectral in vorticity-streamfunction form with 2/3-dealiasing and an integrating-factor RK4 scheme with \( \Delta t=5\times10^{-3} \), integrated to \( t=10 \) on a \( 256\times448 \) grid~\cite{canuto2007Spectral,kassam2005FourthOrdera,orszag1971Eliminationa}. The state stacks \( u \), \( v \), and \( \theta \), so \( N=344064 \).

The varied parameter is \( Re \). Training uses \( Re\in\{250,416.7,583.3,750\} \) with \( 101 \) snapshots each (\( M=404 \)), validation uses \( Re\in\{333.3,666.7\} \) and \( Re=1000 \) with \( 81 \) snapshots each, and the tests use \( Re=561.6 \) (within the training range) and \( Re=1250 \) (outside the training range) with \( 201 \) snapshots each.

Budget selection returns \( P^\star=481.69 \) for \( p=2 \) and \( P^\star=4183.03 \) for \( p=3 \).

\begin{figure}[tb!]
    \centering
    \includegraphics[width=\textwidth]{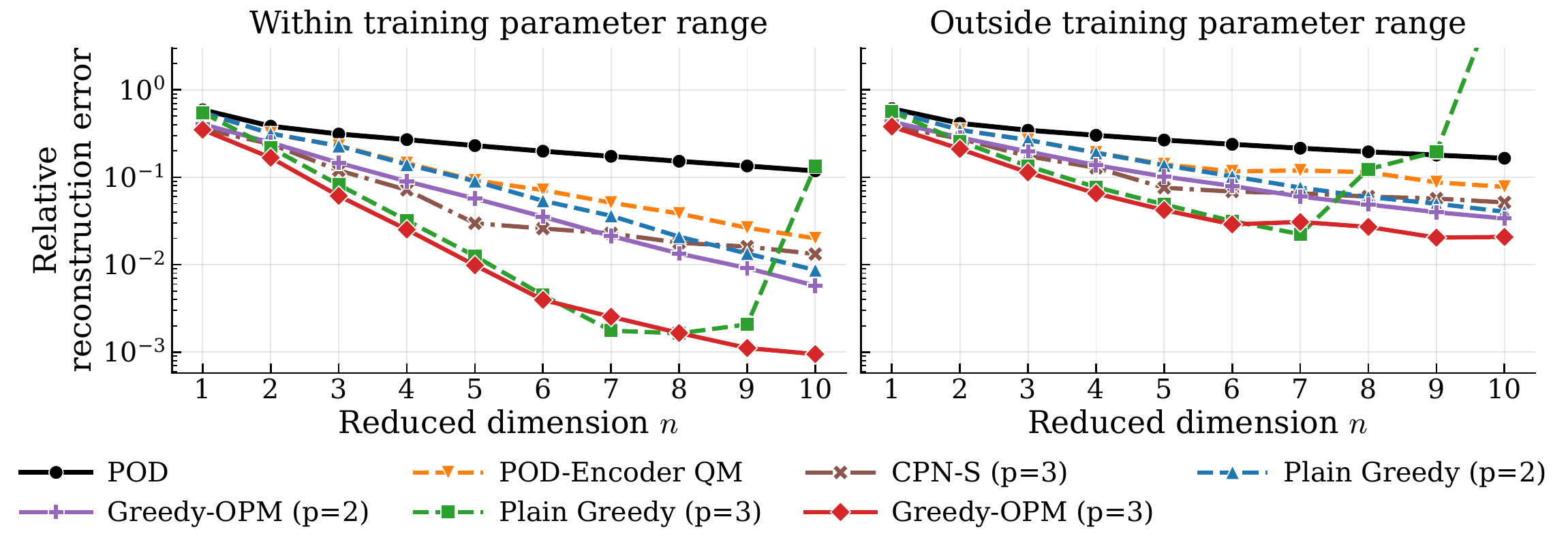}
    \caption{Boussinesq rising thermal plume. Relative reconstruction error~\eqref{eq:rel_err} against \( n \) on the test parameters within (\( Re=561.6 \)) and outside (\( Re=1250 \)) the training range.}\label{fig:bouss_error}
\end{figure}

\begin{figure}[t!]
    \centering
    \includegraphics[width=\textwidth]{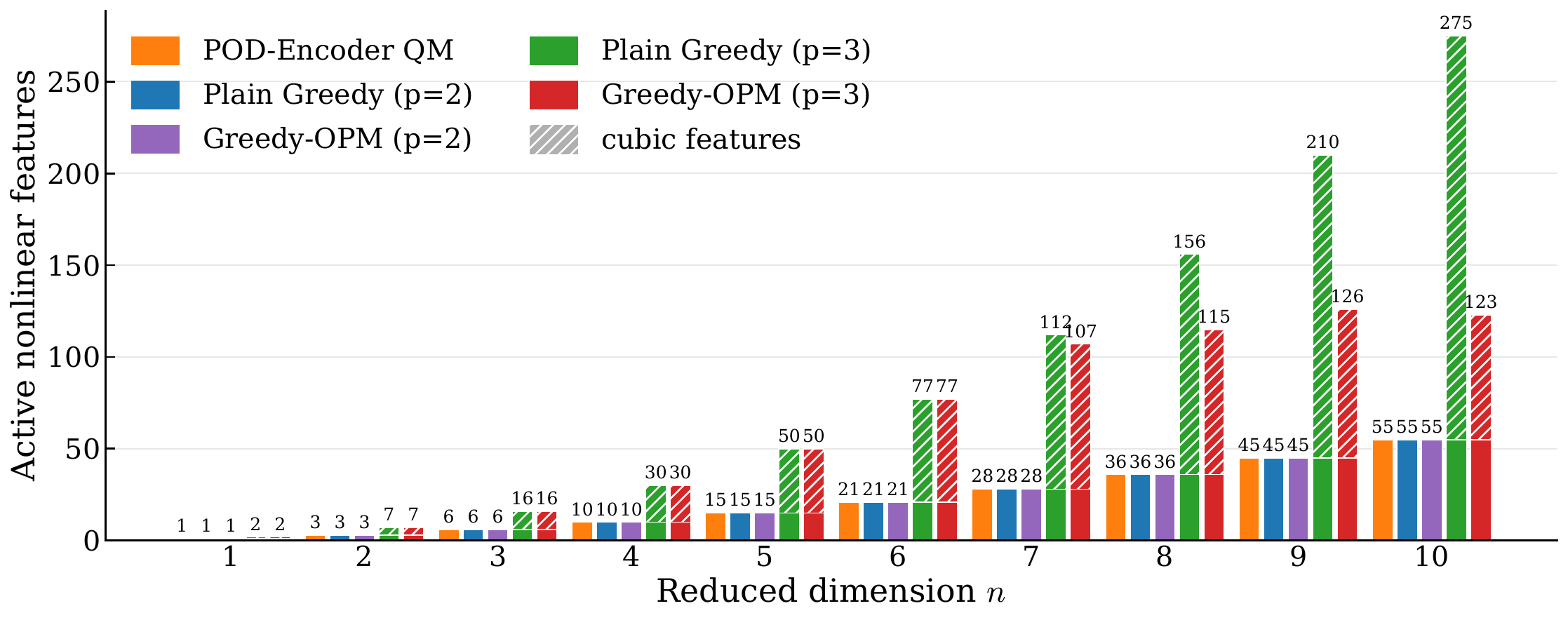}
    \caption{Boussinesq rising thermal plume. Active nonlinear features against \( n \), degree-split for Greedy-OPM \( (p=3) \).}\label{fig:bouss_active}
\end{figure}

\subsubsection{Boussinesq rising thermal plume: Results} \Cref{fig:bouss_error} shows the error against \( n \) for the two test parameters. Within the training range, the Greedy-OPM and Plain Greedy achieve comparable accuracy up to about $n = 7$, when the encoder-decoder pair constructed with Plain Greedy starts to become unstable. 

\begin{figure}[htbp!]
    \centering
    \includegraphics[width=\textwidth]{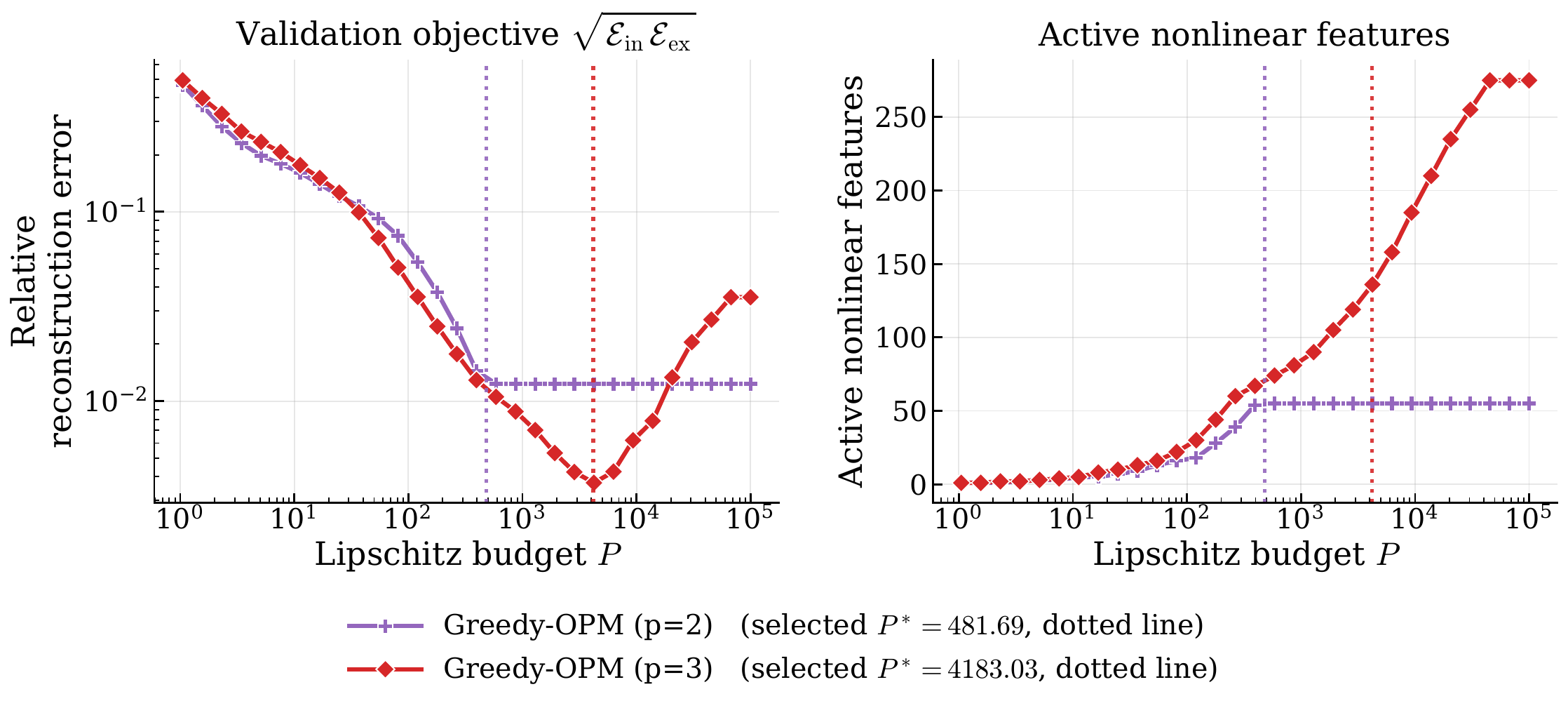}
    \caption{Boussinesq rising thermal plume. Validation objective \( \Ecal_{\mathrm{val}} \) and number of active nonlinear features against the Lipschitz budget \( P \) at \( n=10 \), with the selected \( P^\star \) marked.}\label{fig:bouss_budget}
\end{figure}

\begin{figure}[htbp!]
    \centering
    \includegraphics[width=0.8\linewidth]{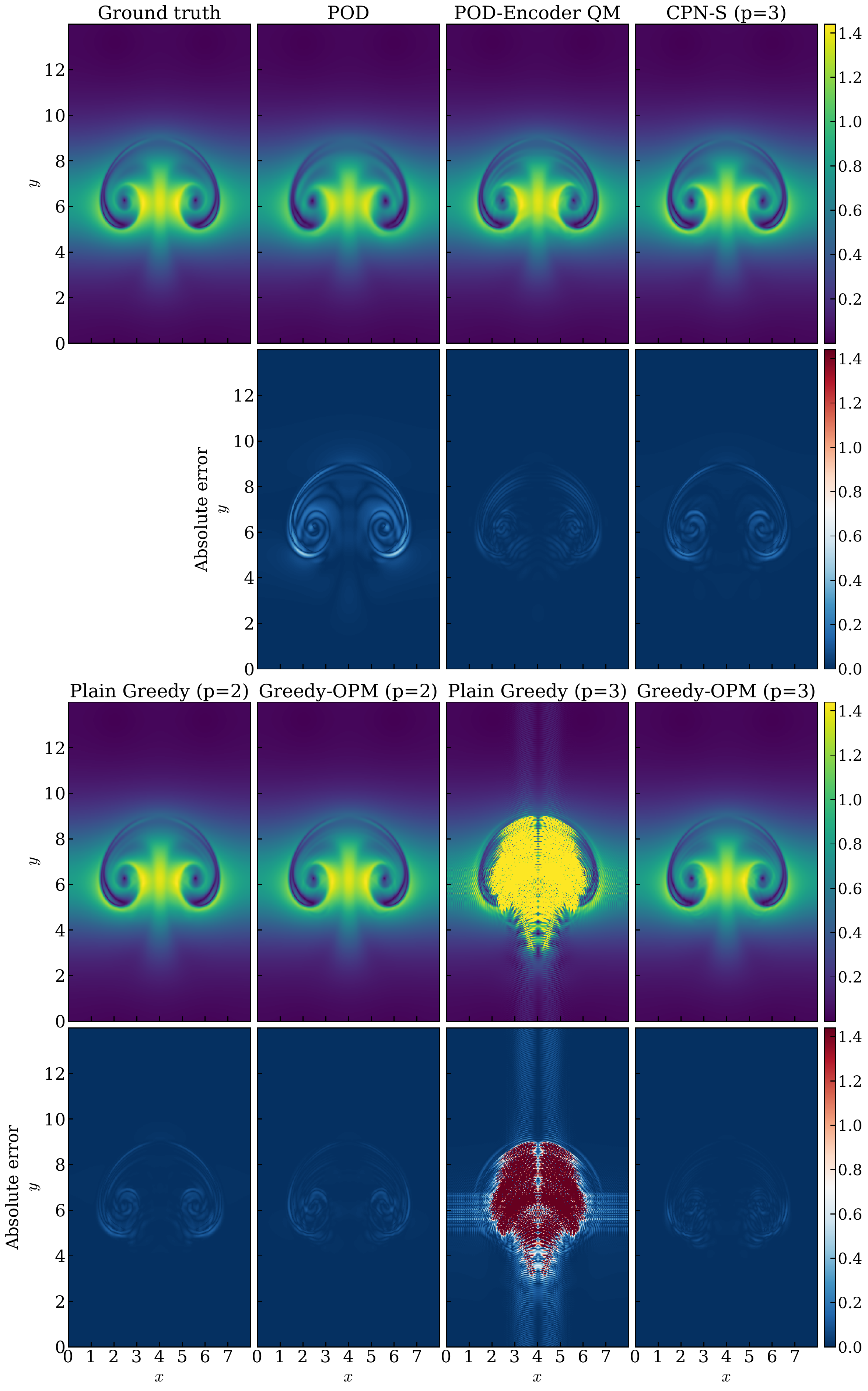}
    \caption{Boussinesq rising thermal plume. Speed \( \sqrt{u^2+v^2} \) at \( t=10 \) for the test parameter outside the training range (\( Re=1250 \)), reconstructed at \( n=10 \), with the absolute error below each reconstruction.}\label{fig:bouss_speed}
\end{figure}

\Cref{fig:bouss_active} shows the active features over the reduced dimensions. Greedy-OPM \( (p=3) \) keeps \( 123 \) of the \( 275 \) candidate features at \( n=10 \), \( 55 \) quadratic and \( 68 \) cubic. From \( n=7 \) on, where the budget constraint becomes active, the size of its active set stays nearly constant while the quadratic share grows, so the cubic features offered by each additional mode are rejected. CPN-S uses \( 144 \) distinct nonlinear basis functions at \( n=10 \). 
We emphasize that Greedy-OPM \( (p=3) \) uses fewer features than the Plain Greedy method and even CPN-S while achieving a competing or lower error, demonstrating how the selection of modes and fitting the coefficient matrix constrained by the Lipschitz budget effectively regularizes the model for stable performance on unseen data.

\Cref{fig:bouss_budget} shows the validation objective over the budget sweep at \( n=10 \). For \( p=3 \) the objective has a distinct minimum at \( P^\star \) and rises by an order of magnitude toward the unconstrained end of the sweep, where all \( 275 \) features are active. For \( p=2 \) the objective saturates once all \( 55 \) quadratic features are active.

\Cref{fig:bouss_speed} shows the speed field at the final time outside the training range. The error of \gls{pod} is concentrated on the rim of the cap and the rolled-up lobes, Plain Greedy \( (p=3) \) breaks down over the whole plume head, with errors exceeding the maximum speed, and Greedy-OPM \( (p=3) \) leaves only thin traces along the rim and inside the lobes.

%% file: sections/conclusion.tex
\section{Conclusion}\label{sec:conclusion}
The contribution of this work is to show that enforcing empirical orthogonality of the polynomial features enables efficient polynomial-manifold learning. When the polynomial features are orthonormal on the training data, their contributions to reconstruction error separate and their contributions to a bound on decoder sensitivity can be quantified individually. This makes it possible to account for accuracy and sensitivity within the construction of the manifold.

Greedy-OPM exploits this structure to select encoder modes according to the reconstruction error achievable under a prescribed sensitivity budget. The resulting constrained fit reduces to a weighted group-$\ell_{2,1}$ projection, so that feature selection and sensitivity control are incorporated directly into the greedy criterion while retaining an explicit Lipschitz bound. The numerical experiments show that this control can prevent the loss of generalization exhibited by polynomial manifolds trained on the reconstruction error only while preserving their approximation gains.

%% file: sections/extra_p5_plots.tex
\FloatBarrier
\section{Extra results for CPN-S (\texorpdfstring{$p=5$}{p=5})}\label{app:extra_plots}

We present additional results for all experiments in \Cref{sec:numerics}, in which we compare our proposed method with the CPN-S method with polynomial degree \( p=5 \). Reconstruction errors are the same as \Cref{fig:lad_error,fig:bur_error,fig:bouss_error} but for CPN-S (\( p = 5 \)).

\begin{figure}[h!]
\includegraphics[width=\textwidth]{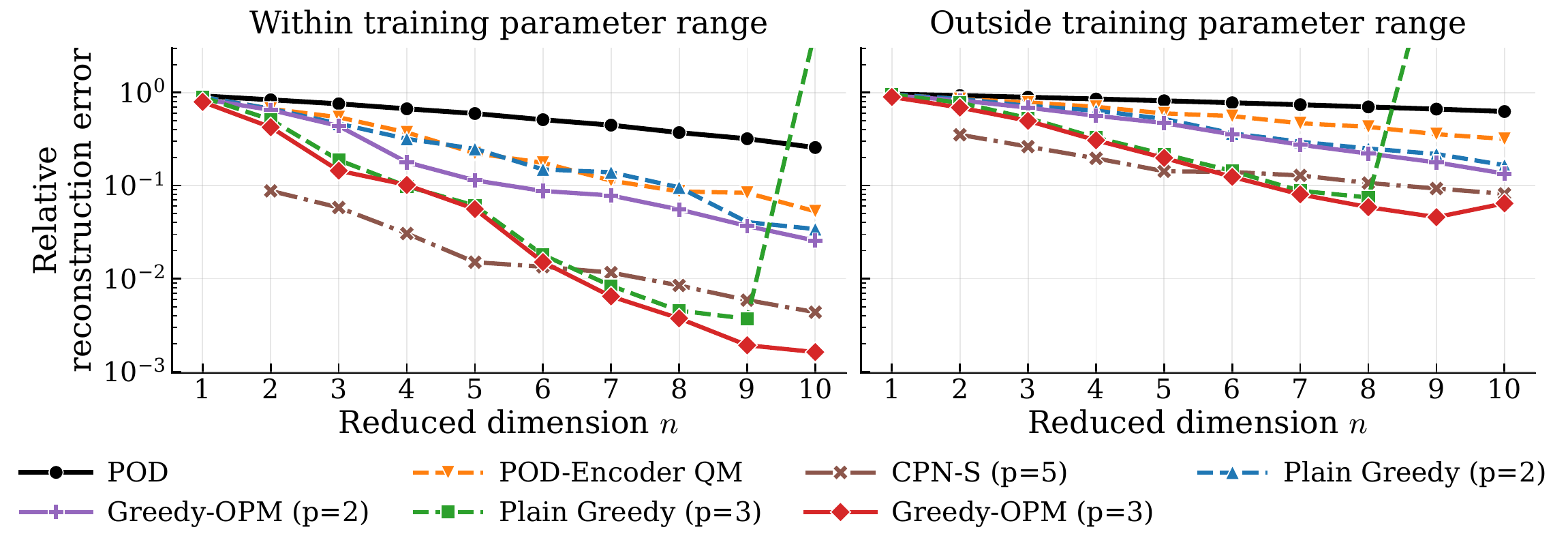}
\caption{Linear advection-diffusion.}
\label{fig:lad_error_p5}
\end{figure}

\begin{figure}[h!]
\includegraphics[width=\textwidth]{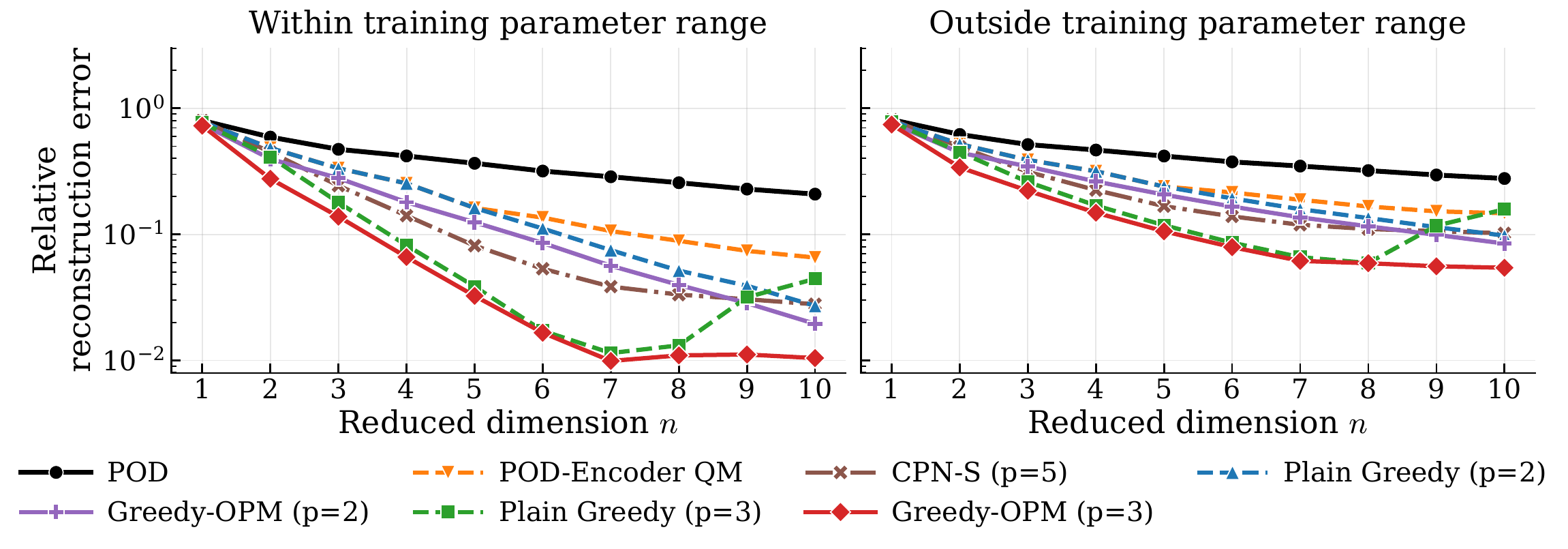}
        \caption{Two-dimensional viscous Burgers.}\label{fig:bur_error_p5}
\end{figure}

\begin{figure}[h!]
\includegraphics[width=\textwidth]{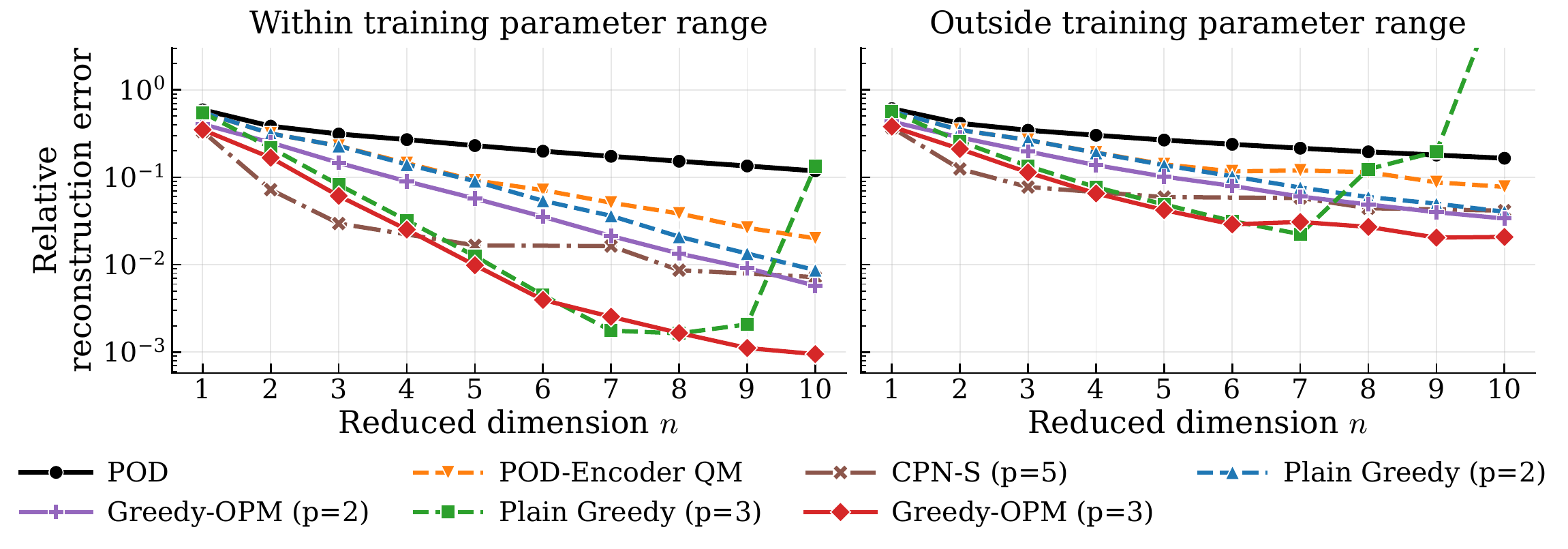}
        \caption{Boussinesq rising thermal plume.}\label{fig:bouss_error_p5}
        \end{figure}